\documentclass[11pt]{article}

\usepackage[final]{acl}

\usepackage{times}
\usepackage{latexsym}

\usepackage[T1]{fontenc}

\usepackage[utf8]{inputenc}

\usepackage{microtype}

\usepackage{inconsolata}
\usepackage{inconsolata}

\usepackage{graphicx}
\usepackage{booktabs}
\usepackage{makecell}
\usepackage{xspace,mfirstuc,tabulary}
\usepackage{tabularray}
\usepackage{tcolorbox}
\UseTblrLibrary{booktabs}
\usepackage{pifont}
\usepackage{amsmath}
\usepackage{subcaption}
\usepackage{xcolor}
\usepackage{amsmath}
\usepackage{tabularx} 
\usepackage{array}     
\usepackage{natbib}
\tcbuselibrary{breakable, skins}
\usepackage[table]{xcolor}
\usepackage{multirow}

\definecolor{actioncolor}{HTML}{2878B5}
\definecolor{mmcolor}{HTML}{8055A6}

\newcommand{\actionlabel}{%
  \textbf{\textcolor{actioncolor}{Action}}%
}
\newcommand{\mmlabel}{%
  \textcolor{mmcolor}{%
    \bfseries\shortstack[l]{Mental\\Model}%
  }%
}
\newcommand{\mmcategory}[1]{%
  \textcolor{mmcolor}{%
    \bfseries\itshape\shortstack[c]{#1}%
  }%
}
\newcommand{\mmcol}[1]{%
  \multicolumn{1}{
    l!{\color{black!35}\vrule width 0.7pt}
    @{\hspace{2pt}}
  }{#1}%
}

\newcommand{\cmark}{\ding{51}}
\newcommand{\xmark}{\ding{55}}
\newcommand{\up}{ \textcolor{green!60!black}{$\Uparrow$}}

\newcommand{\datasetname}{\textsc{Almanac}\xspace}
\title{Humans' \datasetname: A Human Collaboration Dataset of \underline{A}ction-\underline{L}evel \underline{M}ental Model \underline{AN}notations for \underline{A}gent \underline{C}ollaboration}

\author{
 \textbf{Jiaju Chen\textsuperscript{1}},
 \textbf{Yuxuan Lu\textsuperscript{1}},
 \textbf{Jiayi Su\textsuperscript{1}},
 \textbf{Chaoran Chen\textsuperscript{2}},
  \\
 \textbf{Songlin Xiao\textsuperscript{1}},
 \textbf{Zheng Zhang\textsuperscript{3}},
 \textbf{Yun Wang\textsuperscript{4}},
 \textbf{Yunyao Li \textsuperscript{3}},
 \textbf{Jian Zhao\textsuperscript{5}},
   \\
 \textbf{Tongshuang Wu\textsuperscript{6}},
 \textbf{Toby Jia-Jun Li\textsuperscript{2}},
 \textbf{Dakuo Wang\textsuperscript{1}},
 \textbf{Bingsheng Yao\textsuperscript{1}\thanks{~Corresponding author. Contact: \href{b.yao@northeastern.edu}{b.yao@northeastern.edu}, \href{mailto:chen.jiaju2@northeastern.edu}{chen.jiaju2@northeastern.edu}.} },
\\
\\
 \textsuperscript{1}Northeastern University,
 \textsuperscript{2}University of Notre Dame,
 \textsuperscript{3}Adobe,\\
 \textsuperscript{4}Microsoft Research Asia,
 \textsuperscript{5}University of Waterloo,
 \textsuperscript{6}Carnegie Mellon University
}

\begin{document}
\maketitle
\begin{abstract}
Recent advances in LLM agents have enabled complex cognitive capabilities, such as multi-step reasoning, planning, and tool use, that increasingly position these agents as human collaborators. 
Effective collaboration, however, requires collaborators to continuously maintain and align mental models of their own reasoning, partners' intentions, and shared goals during the collaborative process. 
Today's agents rarely develop such capabilities since they are primarily optimized for task completion, and the community lacks authentic human collaboration data with action-level mental model annotations that could guide agents toward process-level collaborative competence.
To bridge this gap, we present \textbf{\datasetname}\footnote{\datasetname is available at \url{https://huggingface.co/datasets/NEU-HAI/Almanac}.}, a dataset of \textbf{\underline{A}}ction-\textbf{\underline{L}}evel \textbf{\underline{M}}ental model \textbf{\underline{AN}}notations for \textbf{\underline{A}}gent \textbf{\underline{C}}ollaboration built from the Map Task, a classic dyadic routing task from social science. 
\datasetname contains 2,987 collaboration actions, each paired with theory-informed mental model annotations that record the participants' self-reasoning, perceived partner intent, and perceived team goal.
We benchmark six LLMs on predicting humans' next-turn behavior and mental models. 
Our results demonstrate \datasetname's utility in evaluating models' ability to simulate human collaborative behaviors and infer their underlying mental models. The project page is available at \url{https://human-agent-collab.hailab.io/almanac}.
\end{abstract}

\section{Introduction}

\begin{figure}[t]
    \centering
    \includegraphics[width=1.00\linewidth]{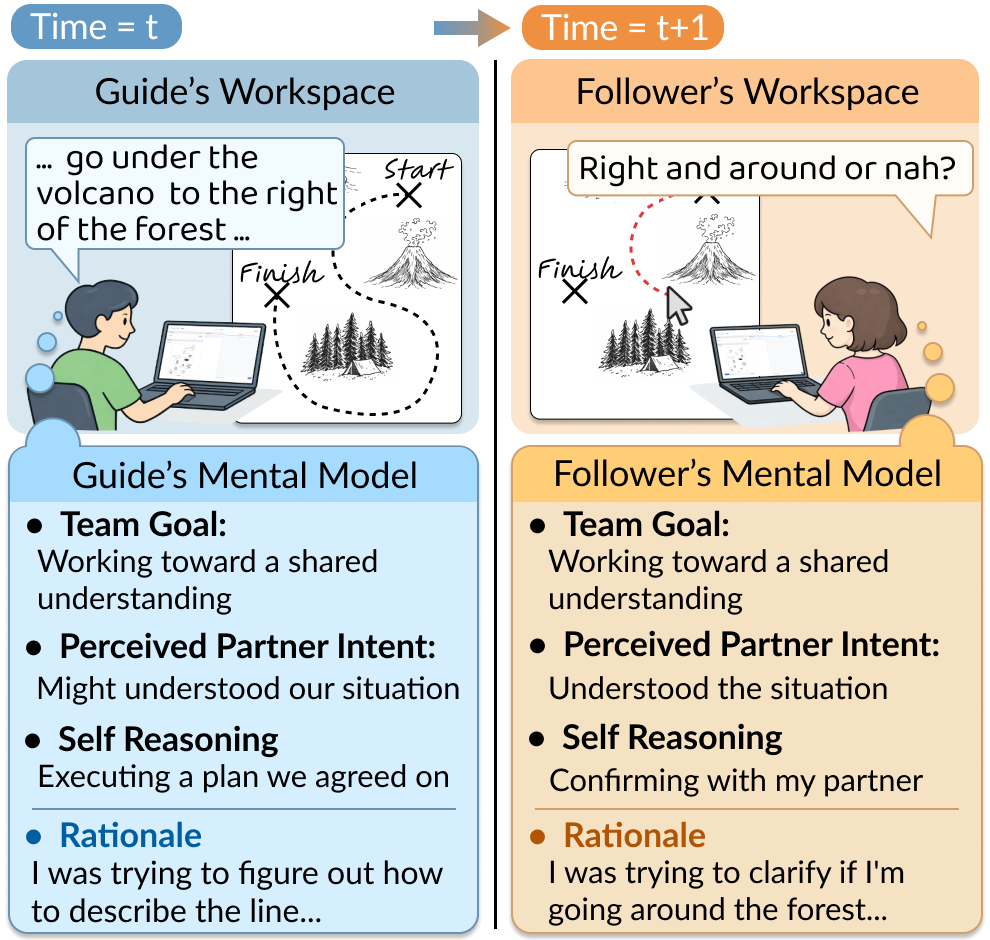}
    \caption{A sample datapoint of \datasetname, which contains participants' actions, mental models (team goal, perceived partner intent, self-reasoning), and a free-form rationale. 
    We implement the Map Task, a classic dyadic routing task, to collect human collaborative behaviors and action-level mental model annotations.}
    \label{fig:maptask_workflow}
\end{figure}

Recent advances in Large Language Model (LLM) agents have enabled complex cognitive capabilities for task-solving (i.e., multi-step reasoning, planning, tool use, and behavioral modeling) that increasingly position these agents as collaborative partners in human workflows \cite{park2024generative, Singh2025AgenticRA, chen-etal-2026-multi-agent,luUXAgentSystemSimulating2025a,luAgentAutomatedScalable2026,yaoDPRFGeneralizableDynamic2025a}. 
A growing body of work designs LLM agents for complex collaborative tasks such as programming and collaborative writing \cite{he2025llm, venkatraman2025collabstory}, where agents engage in multi-turn communication and coordination with humans. 
In practice, these agents resemble remote human collaborators in important ways, since both operate through structured, text-based channels and lack the non-verbal cues present in face-to-face interactions \cite{olson2000distance, yao2025through}. 
Grounding human-agent collaboration in this analogy allows researchers to draw on decades of research on how remote human collaborators build trust, maintain awareness, and coordinate effectively \cite{clark1991grounding, gutwin2002descriptive}, and whether established principles of remote human collaboration persist or break down when partners are LLM agents.

Effective collaboration, however, requires a distinct set of capabilities that task-solving proficiency alone does not provide. 
Research on human-human collaboration~\cite{cannon1993shared} has established that successful collaboration depends on collaborators' ability to continuously maintain and align mental models during the collaborative process, including self-reasoning about their own actions, perceived partner intent, and understanding of the shared team goal~\cite{malone1994interdisciplinary, gutwin2002descriptive, marks2001temporally}. 
The cognitive effort involved in aligning these mental models is what enables collaborators to coordinate actions, establish mutual understanding, and resolve misalignment over time. 

\begin{table*}[ht!]
    \centering
    \resizebox{1.00\linewidth}{!}{
    \begin{booktabs}{
    colspec={lQ[l,4cm]Q[l,6.5cm]ccc},
    row{1}={font=\bfseries\small},
    width=\linewidth,
    column{1,2,3}={halign=l}
}
\toprule[1pt]
Dataset & Interaction Type & Scenario & Mental Model Annotation & Real Human Data \\
\midrule
Deal or No Deal & Human--Human & Negotiation & \xmark & \cmark \\
Mutual Friends & Human--Human & Information Sharing & \xmark & \cmark \\
CaSiNo & Human--Human & Negotiation & \xmark & \cmark \\
\midrule
Rifts & Human--Agent & Dialogue Clarification \& Grounding & \xmark & \cmark \\
CoGym & Human--Agent & Multi-Task Collaboration & \xmark & \cmark \\
\midrule
ToolBench & Single Agent & Tool Use & \xmark & \xmark \\
WebArena & Single Agent & Web Navigation & \xmark & \xmark \\
$\tau$-Bench & Single Agent & Agent-User-Tool Interaction & \xmark & \xmark \\
Multi-Agent-Bench & Multi-Agent & Multi-Task Coordination & \xmark & \xmark \\
\midrule
\datasetname & Human--Human & Collaborative Routing & \cmark & \cmark \\
\bottomrule[1pt]
\end{booktabs}}
\caption{Properties of existing representative datasets compared to \datasetname. }
\label{tab:dataset_comparison}
\end{table*}

However, most current human-agent collaboration remains focused on task-directed exchanges in which the human issues instructions and the agent responds with actions such as tool calls or information retrieval \cite{qi2025agentif}. 
LLM agents in such settings are often optimized for independent task completion rather than for maintaining the mental models needed for effective collaboration.
Existing agent benchmarks such as ToolBench \cite{qin2023toolllm}, WebArena \cite{Zhou2023WebArenaAR}, $\tau$-Bench \cite{yao2024tau}, and MultiAgentBench \cite{zhu-etal-2025-multiagentbench} evaluate whether agents can complete tasks under instructions or coordinate task execution, not whether their behaviors support effective collaboration with humans. 
Thus, agents are rarely exposed to the interaction patterns and cognitive processes that characterize successful collaboration. 
Existing human collaboration datasets \cite{lewis-etal-2017-deal, chawla-etal-2021-casino} reinforce this gap by capturing observable interaction content, such as dialogues and outcomes, while omitting the critical cognitive content that underlies collaborative behaviors (e.g., collaborators' mental models). 
To our knowledge, no dataset \textbf{pairs human collaboration behaviors with action-level mental model annotations grounded in collaboration theory} that can guide agents toward collaborative competence.

In this work, we present \textbf{\datasetname}, a dataset of \textbf{\underline{A}}ction-\textbf{\underline{L}}evel \textbf{\underline{M}}ental model \textbf{\underline{AN}}notations for \textbf{\underline{A}}gent \textbf{\underline{C}}ollaboration built from the Map Task \cite{anderson1991hcrc}, a classic dyadic routing task from social science in which two participants collaborate to reproduce a route through text-based communication and workspace actions such as drawing. 
We implement the Map Task on a configurable research platform \cite{yao2025through} and develop an annotation framework grounded in collaboration theories \cite{cannon1993shared, marks2001temporally, gutwin2002descriptive} to capture participants' mental models at the action level. 
\datasetname contains 2,987 collaboration actions from 50 participants across 25 dyadic sessions, each paired with the participant's own mental model annotation, capturing their self-reasoning, perceived partner intent, and perceived team goal, along with a free-form rationale explaining the cognitive process behind the action. 
We benchmark \textbf{six} state-of-the-art LLMs under prompt-based and fine-tuning settings on two complementary tasks: next-turn behavior prediction and mental model prediction. 
Results show that mental model annotations provide useful signals for predicting human collaborative behavior, but current LLMs remain limited in inferring humans' internal reasoning.

Our contributions are as follows. 
First, we collect \datasetname, the first human collaboration dataset that pairs authentic collaborative behaviors with theory-informed, action-level mental model annotations. 
Second, we design a theory-informed annotation framework that combines in-session checkpoints with post-session retrospective labeling to capture collaborators' action-level mental models and free-form rationales. 
Third, we benchmark six LLMs and show that mental models offer useful signals for modeling collaborative behavior.

\section{Related Work}

\subsection{Collaboration Datasets and Benchmarks}

Existing collaboration-related datasets and benchmarks fall into three categories based on the collaborators involved, as shown in Table \ref{tab:dataset_comparison}: human-human, human-agent, and agent benchmarks. 
The first category consists of datasets grounded in social experiments with humans and mainly designed for dialogue modeling, such as DealNoDeal \cite{lewis-etal-2017-deal}, MutualFriends \cite{he-etal-2017-learning}, and CaSiNo \cite{chawla-etal-2021-casino}.
The second category focuses on human-agent collaboration, where recent benchmarks measure LLMs' grounding behaviors during human-LLM interaction \cite{shaikh-etal-2025-navigating, poelitz2026benchmark}.
Although both involve human participants, they generally lack annotations of collaborators' mental models \cite{berretta2023defining}, offering limited support for modeling the underlying reasoning that facilitates effective collaboration.

The third category is agent benchmarks, for instance,
ToolBench \cite{qin2023toolllm}, WebArena \cite{Zhou2023WebArenaAR}, and $\tau$-Bench \cite{yao2024tau} assess whether agents can follow high-level instructions and execute multi-step actions in interactive tool-use environments. 
In multi-agent settings, MultiAgentBench \cite{zhu-etal-2025-multiagentbench} evaluates agent performance on tasks such as coding and database error analysis. SOTOPIA \cite{zhou2024sotopia} provides open-ended agent social scenarios and evaluates agents' social intelligence.
However, these benchmarks primarily assess agents' task-solving capabilities and overlook their ability and limitations to coordinate effectively with humans.

\subsection{LLM Agents in Human-Agent Collaboration}
Recent advances have moved LLM agents beyond static text generation toward interactive task execution, expanding their capabilities in two directions.
First, LLM agents can perform increasingly complex tasks that require multi-step planning\cite{yao2022react}, tool invocation\cite{schick2023toolformer, yao2024tau,wangTrajectory2taskTrainingRobust2026,luFireflyIlluminatingLargeScale2026}, and long-horizon interaction \cite{park2024generative, xu2025mem, chen2026collabsim}. 
Second, LLM agents have shown emerging cognitive capabilities relevant to collaboration, including language understanding and natural communication \cite{wang2024survey}, context perception and situational reasoning \cite{yao2022react, chen-etal-2024-storysparkqa}, and behavior modeling based on provided personas or backstories \cite{luUXAgentSystemSimulating2025a,luAgentAutomatedScalable2026,yaoDPRFGeneralizableDynamic2025a}. To further align agent behaviors with human expectations, recent work has explored supervised fine-tuning on human demonstration data \cite{xia-etal-2025-agentrm, wu2025collabllm} and reinforcement learning from human or environment feedback \cite{abdulhai2025consistently, du-etal-2025-simvbg,zhangShopr1RewardingLlms2026,wangCustomerR1PersonalizedSimulation2025a,luWEBSERVBrowserServerEnvironment2025}.

These advances have motivated the use of LLM agents in various human workflows \cite{xiao2024flowbench, sun2024explore, shihab2025effects, chen2025character, AIMeeting,yangRECOVERDesigningLarge2026a,sunLLMAgentMeets2025a}. However, in many current human-agent collaborations, LLM agents act as assistive systems that respond to human instructions, rather than as equal collaborators that actively infer human partners' intents and evolving mental states throughout the collaboration \cite{chen2025need}.

\begin{figure*}
    \centering
    \includegraphics[width=1.00\linewidth]{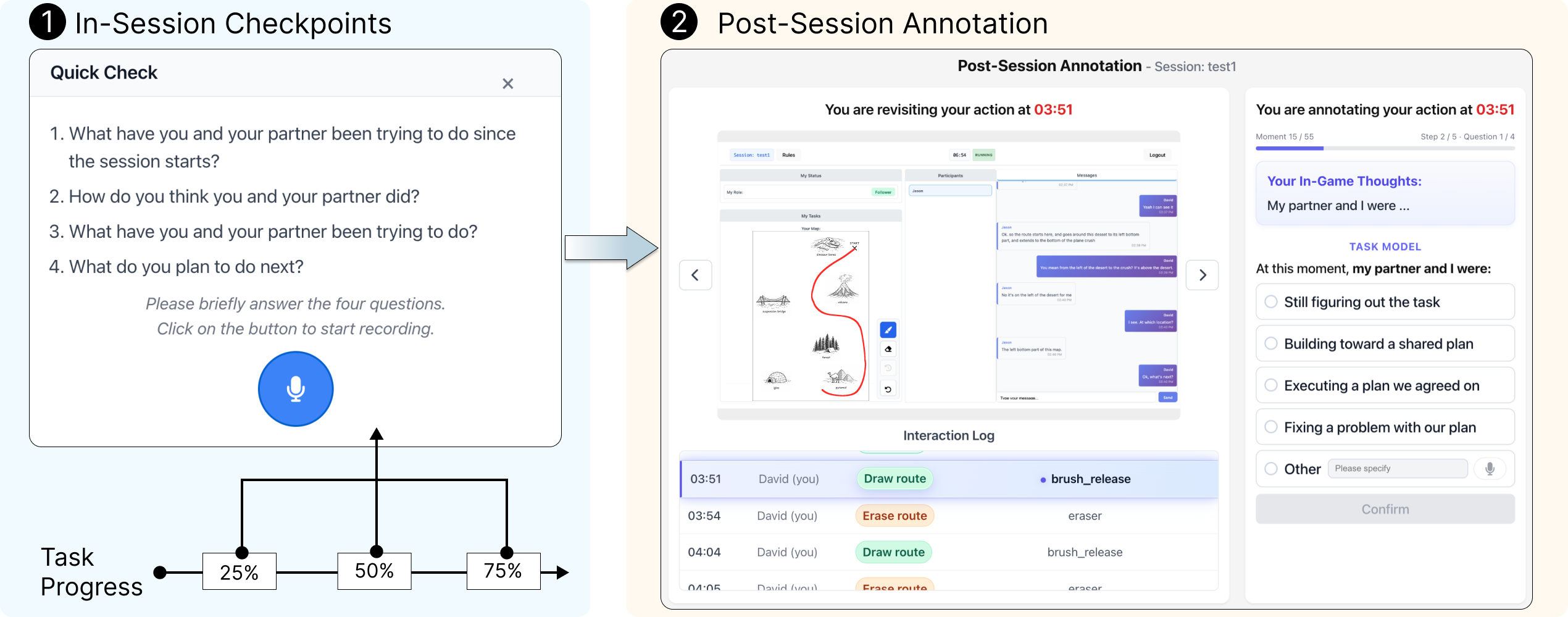}
    \caption{Annotation workflow and interfaces of \datasetname. Participants first complete the Map Task while providing brief in-session mental model annotations at checkpoints (25\%, 50\%, 75\%). Afterward, they review their action trajectory to retrospectively annotate the team goal, self-reasoning, and perceived partner intent per action.}
    \label{fig:annotation}
\end{figure*}

\section{\datasetname}

We describe the design and construction of \datasetname in three parts: the annotation framework (Sec. \ref{sec:annotation}), the data collection process (Sec. \ref{sec:collection}), and the resulting dataset details (Sec. \ref{sec:dataset}).


\subsection{Annotation Framework}
\label{sec:annotation}

To collect humans' mental models during collaboration, we design a two-step annotation framework. 
The \textit{in-session} annotation elicits participants' real-time mental models at key moments in the interaction, while the \textit{post-session} annotation uses the in-session annotation as memory anchors to support action-level mental model annotation.

Grounded in theories of teamwork process~\cite{marks2001temporally}, situation awareness~\cite{endsley2017direct}, common ground~\cite{traum1995computational}, and workspace awareness~\cite{gutwin2002descriptive}, we translate mental models into three action-level components that are both theoretically central and practically elicitable in the Map Task: the participant's self-reasoning about their own actions, perceived partner intent, and understanding of the shared team goal. 
In addition to these structured components, participants provide a free-form rationale explaining each action (See Appendix \ref{app:schema}).

\paragraph{Step 1: In-Session Annotation}

During the Map Task, we periodically elicit participants' mental states through brief in-session checkpoints at 25\%, 50\%, and 75\% of the route drawing progress. 
We selected three checkpoints spaced at quarter intervals to capture the evolution of mental models across early, middle, and late task stages while keeping interruptions to a minimum~\cite{endsley2017direct, schinkel2024implementing}. 
Specifically, we implement a rule-based mechanism that tracks participants’ key actions (e.g., sending a message or drawing a route). 
At each checkpoint (left part of Figure \ref{fig:annotation}), the system asks participants to briefly report their perceived team goal, partner's intention, and self-reasoning since the last checkpoint. 
To reduce participant burden, responses are collected via voice recordings and automatically transcribed. Each checkpoint typically takes 10--20 seconds.

\paragraph{Step 2: Post-Session Annotation}

Immediately after task completion, participants retrospectively annotate their action-level mental models. 
To support recall, the annotation interface (right part of Figure \ref{fig:annotation}) presents (1) the participant’s action trajectory, (2) the screenshot of each action, and (3) the temporally closest in-session response, which serves as a memory anchor \cite{simulatedrecall} for reconstructing reasoning around that moment. 
For each action, participants first articulate their action rationale through voice recording. 
Then, they complete four single-choice questions that capture their mental models (i.e., self-reasoning, perceived partner intent, and perceived team goal, along with an additional item indicating perceived alignment).
We additionally validate the retrospective post-session annotation, as reported in Appendix \ref{app:mm_annotation_validation}).

\begin{table*}[t]
\centering
\small
\setlength{\tabcolsep}{3.8pt}
\renewcommand{\arraystretch}{1.08}

\resizebox{\textwidth}{!}{%
\begin{tabular}{
  @{}l c
>{\raggedright\arraybackslash}l|
  rrrr|rrrr|rrrr
  @{}
}
\toprule

\multicolumn{2}{c}{%
  \multirow[c]{4}{*}{\textbf{Category}}%
}
& \multirow[c]{4}{*}{\textbf{\# Dimension / Session}}
& \multicolumn{4}{c}{\textbf{All}}
& \multicolumn{4}{c}{$\boldsymbol{C_{not\_visible}}$}
& \multicolumn{4}{c}{$\boldsymbol{C_{visible}}$} \\

\multicolumn{2}{c}{} & {}
& \multicolumn{4}{c}{\textit{25 sessions, 2987 actions}}
& \multicolumn{4}{c}{\textit{12 sessions, 1469 actions}}
& \multicolumn{4}{c}{\textit{13 sessions, 1518 actions}} \\

\cmidrule(lr){4-7}
\cmidrule(lr){8-11}
\cmidrule(lr){12-15}

\multicolumn{2}{c}{} & {}
& \multicolumn{2}{c}{Train}
& \multicolumn{2}{c}{Test}
& \multicolumn{2}{c}{Train}
& \multicolumn{2}{c}{Test}
& \multicolumn{2}{c}{Train}
& \multicolumn{2}{c}{Test} \\

\multicolumn{2}{c}{} & {}
& Avg & SD & Avg & SD
& Avg & SD & Avg & SD
& Avg & SD & Avg & SD \\

\midrule

\multicolumn{2}{c}{\multirow{6}{*}{\actionlabel}}
& Total actions
& 117.7 & 87.7 & 125.2 & 84.5
& 123.6 & 96.4 & 119.0 & 99.2
& 112.4 & 84.0 & 131.3 & 88.9 \\

\multicolumn{2}{c}{}
& Message
& 74.0 & 57.1 & 74.3 & 47.0
& 88.9 & 75.0 & 67.0 & 45.9
& 60.6 & 33.1 & 81.7 & 57.1 \\

\multicolumn{2}{c}{}
& Draw
& 27.4 & 31.2 & 25.3 & 16.2
& 19.4 & 11.7 & 22.7 & 18.0
& 34.5 & 41.3 & 28.0 & 17.6 \\

\multicolumn{2}{c}{}
& Erase
& 8.8 & 9.0 & 17.2 & 21.9
& 7.9 & 8.3 & 24.7 & 30.6
& 9.7 & 9.9 & 9.7 & 9.5 \\

\multicolumn{2}{c}{}
& Undo
& 6.4 & 7.8 & 7.8 & 8.1
& 6.2 & 7.7 & 3.7 & 5.5
& 6.5 & 8.3 & 12.0 & 9.0 \\

\multicolumn{2}{c}{}
& Reset
& 1.1 & 1.8 & 0.5 & 0.5
& 1.1 & 1.4 & 1.0 & 0.0
& 1.1 & 2.2 & 0.0 & 0.0 \\

\midrule

\mmcol{\multirow{15}{*}{\mmlabel}}
& \multirow[c]{4}{*}{\mmcategory{Team\\goal}}
& \textit{Clear on what to do}
& 43.9 & 19.7 & 56.7 & 22.1
& 40.2 & 12.3 & 56.8 & 16.2
& 47.2 & 24.8 & 56.6 & 31.0 \\

\mmcol{} &
& \textit{Working toward shared understanding}
& 38.7 & 18.3 & 28.9 & 15.1
& 42.1 & 15.6 & 25.7 & 11.4
& 35.6 & 20.7 & 32.1 & 20.1 \\
\mmcol{}
&
& \textit{Unclear, working it out}
& 11.8 & 12.0 & 9.8 & 8.5
& 12.2 & 7.4 & 11.9 & 7.2
& 11.4 & 15.5 & 7.7 & 10.6 \\
\mmcol{}
&
& \textit{Still figuring out task}
& 4.8 & 3.9 & 3.4 & 3.8
& 5.0 & 4.2 & 3.5 & 5.7
& 4.5 & 3.9 & 3.3 & 2.0 \\

\cmidrule(lr){2-15}

\mmcol{}
& \multirow[c]{5}{*}{\mmcategory{Partner\\intent}}
& \textit{Understood and aligned}
& 57.9 & 24.0 & 65.4 & 19.7
& 49.1 & 22.9 & 59.8 & 21.6
& 65.7 & 23.2 & 70.9 & 20.3 \\
\mmcol{}
&
& \textit{Probably understood}
& 19.5 & 16.9 & 15.1 & 14.0
& 24.2 & 16.2 & 19.8 & 17.1
& 15.4 & 17.3 & 10.4 & 11.5 \\
\mmcol{}
&
& \textit{Waiting for more information}
& 16.1 & 15.6 & 14.9 & 6.5
& 19.2 & 19.9 & 16.1 & 8.3
& 13.3 & 10.8 & 13.7 & 5.5 \\
\mmcol{}
&
& \textit{Misunderstood or misaligned}
& 3.7 & 4.0 & 3.3 & 3.9
& 4.4 & 3.6 & 3.0 & 4.9
& 3.0 & 4.4 & 3.5 & 3.7 \\
\mmcol{}
&
& \textit{No clear signal}
& 1.7 & 3.5 & 1.3 & 2.1
& 1.3 & 2.7 & 1.3 & 2.2
& 2.1 & 4.2 & 1.4 & 2.5 \\

\cmidrule(lr){2-15}

\mmcol{}
& \multirow[c]{6}{*}{\mmcategory{Self\\reasoning}}
& \textit{Executing agreed plan}
& 41.8 & 25.2 & 41.5 & 11.3
& 33.9 & 25.5 & 35.6 & 13.2
& 49.0 & 23.9 & 47.4 & 6.1 \\
\mmcol{}
&
& \textit{Confirming with partner}
& 27.9 & 15.9 & 20.3 & 7.9
& 30.3 & 19.3 & 21.2 & 12.4
& 25.8 & 12.7 & 19.3 & 1.2 \\
\mmcol{}
&
& \textit{Repairing mistake or misunderstanding}
& 11.9 & 10.9 & 12.4 & 10.1
& 10.6 & 8.3 & 14.4 & 14.4
& 13.0 & 13.1 & 10.4 & 5.9 \\
\mmcol{}
&
& \textit{Grounding by sharing/requesting info}
& 10.4 & 8.9 & 14.2 & 8.7
& 14.4 & 8.4 & 15.3 & 12.3
& 6.7 & 8.1 & 13.1 & 5.8 \\
\mmcol{}
&
& \textit{Exploring independently}
& 3.8 & 4.9 & 7.7 & 5.7
& 4.6 & 5.1 & 11.1 & 6.5
& 3.1 & 4.9 & 4.4 & 2.0 \\
\mmcol{}
&
& \textit{Waiting for more information}
& 2.1 & 2.8 & 3.2 & 3.6
& 3.5 & 3.4 & 2.5 & 3.5
& 0.8 & 1.2 & 4.0 & 4.3 \\

\bottomrule
\end{tabular}%
}

\caption{Dataset statistics and mental model label distributions of
\datasetname{} across splits and conditions. Mental model values are
reported as session-level averages and standard deviations.}
\label{tab:stats}
\end{table*}

\subsection{\datasetname Data Collection}
\label{sec:collection}

After IRB approval, we recruit a total of 50 participants through snowball sampling \cite{goodman1961snowball}. Participants were paired into 25 dyads, each completing one collaboration session.
We introduce the data collection and curation process hereinafter.

\subsubsection{Participant Background Collection}
With participants' consent, we collected their persona information
through a structured online survey (see Appendix \ref{app: survey}).
The survey contains two sections: \textit{demographic information}, including age, gender, education level, and \textit{collaboration tendency}. 
Collaboration tendency is measured using the TeamQ instrument \cite{britton2017assessing}, a validated scale for capturing participants' attitudes and behavioral tendencies toward collaboration, communication, and collective problem-solving.

\subsubsection{Data Collection Process}
In the Map Task \cite{anderson1991hcrc}, participants are either assigned as the \textit{Guide} or the \textit{Follower}. 
The Guide has a map containing both landmarks and a designated route, while the Follower's map contains only the landmarks. 
The participants need to collaborate to reproduce the route on the follower's map as accurately as possible.

We implement the Map Task following the original protocol \cite{anderson1991hcrc} on a web-based research platform \cite{yao2025through}, which enables remote data collection. 
Participants can communicate via a text-based chat interface. 
For the Follower, the platform offers a set of tools for drawing routes, including a brush, eraser, undo, and reset buttons.
To diversify collaboration behaviors and mental model states, we varied task difficulty by manipulating \textbf{whether the Guide could view the Follower's real-time drawing canvas} as a between-subjects factor. 
In the $C_{visible}$ condition, the Guide's interface displayed both the Guide's map and a live view of the Follower's canvas. 
In the $C_{not\_visible}$ condition, the Guide could only view the Guide's own map. 
Participant pairs were randomly assigned to one of these two conditions.

Prior to the task, participants were walked through an onboarding procedure covering task rules and platform use to ensure they were familiar with the interface before the session began.
In particular, we imposed no time limit on task completion to avoid inducing time-pressure effects that could alter participants' natural collaborative behavior. 
Sessions lasted an average of 28.25 minutes (SD = 15.59)
Participants each received a \$25 Amazon gift card as compensation upon completion.

\subsubsection{Post-Processing}

We remove all personally identifiable information (e.g., names) from the collected data. 
To enable consistent map representation for LLMs' downstream modeling, we standardize all maps to a discrete grid, aligning spatial elements (e.g., landmarks and routes) with grid coordinates. 
The Follower’s drawing trajectory is converted into cell-level representations by marking the cells traversed by the route. 
The text-encodable format helps focus the evaluation on collaboration behavior simulation rather than on LLM agents' image comprehension capabilities.
The map materials and standardization details are reported in Appendix \ref{app:map}.

\subsection{Dataset Details}
\label{sec:dataset}

\subsubsection{Data Structure}

Each session $s$ consists of two participants' personas, action traces, and mental model annotations. 
Each action at time $t$ denoted as $a_t \in \mathcal{A}$ is timestamped and paired with a post-hoc mental 
model annotation $m_t = (r_t, g_t, i_t, e_t, \alpha_t)$, where
$\mathcal{A} = \{\texttt{message}, \texttt{draw}, \texttt{erase}, \texttt{undo}, 
\texttt{reset}\}$ denotes the action space. 
In the mental model tuple, $r_t$ is a free-form rationale , and $g_t$, $i_t$, and $e_t$ are text labels capturing \textit{team goal}, \textit{partner intent}, and 
\textit{self-reasoning} respectively. $\alpha_t$ denotes human annotated \textit{alignment status}.
For each drawing-related action $a_t \in \mathcal{A} \setminus \{\texttt{message}\}$, we record the Follower's canvas state 
$x_t \in \{0,1\}^{H \times W}$ at time $t$. 
Each entry $x_t^{(i,j)}$ indicates whether the corresponding grid cell has been traversed by the route.

\subsubsection{Statistics and Analysis of \datasetname}

\paragraph{Dataset Overview.}
\datasetname contains 25 sessions (12 in $C_{not\_visible}$, 13 in $C_{visible}$), 2,987 human actions with mental model annotations. 
Table \ref{tab:stats} presents the core statistics of \datasetname. Additional partition and mental model distribution details are reported in Appendix \ref{app:distribution}.

We measure task success as the proportion of cells in the follower’s drawing that overlap with the ground-truth route (see Section \ref{sec:eval}). 
Across all sessions, participants achieved an average final accuracy of 0.66 (SD = 0.12), with $C_{not\_visible}$ at 0.67 (SD = 0.12) and $C_{visible}$ at 0.65 (SD = 0.12). 
Sessions in $C_{not\_visible}$ involved more actions on average (122.42, SD = 92.48) than in $C_{visible}$ (116.77, SD = 81.69), reflecting the additional effort required when the Guide could not directly see the Follower’s canvas. 
The large variation in action counts across both conditions suggests substantial differences in teams' collaboration styles. 

\begin{figure}[h]
    \centering
    \includegraphics[width=1.00\linewidth]{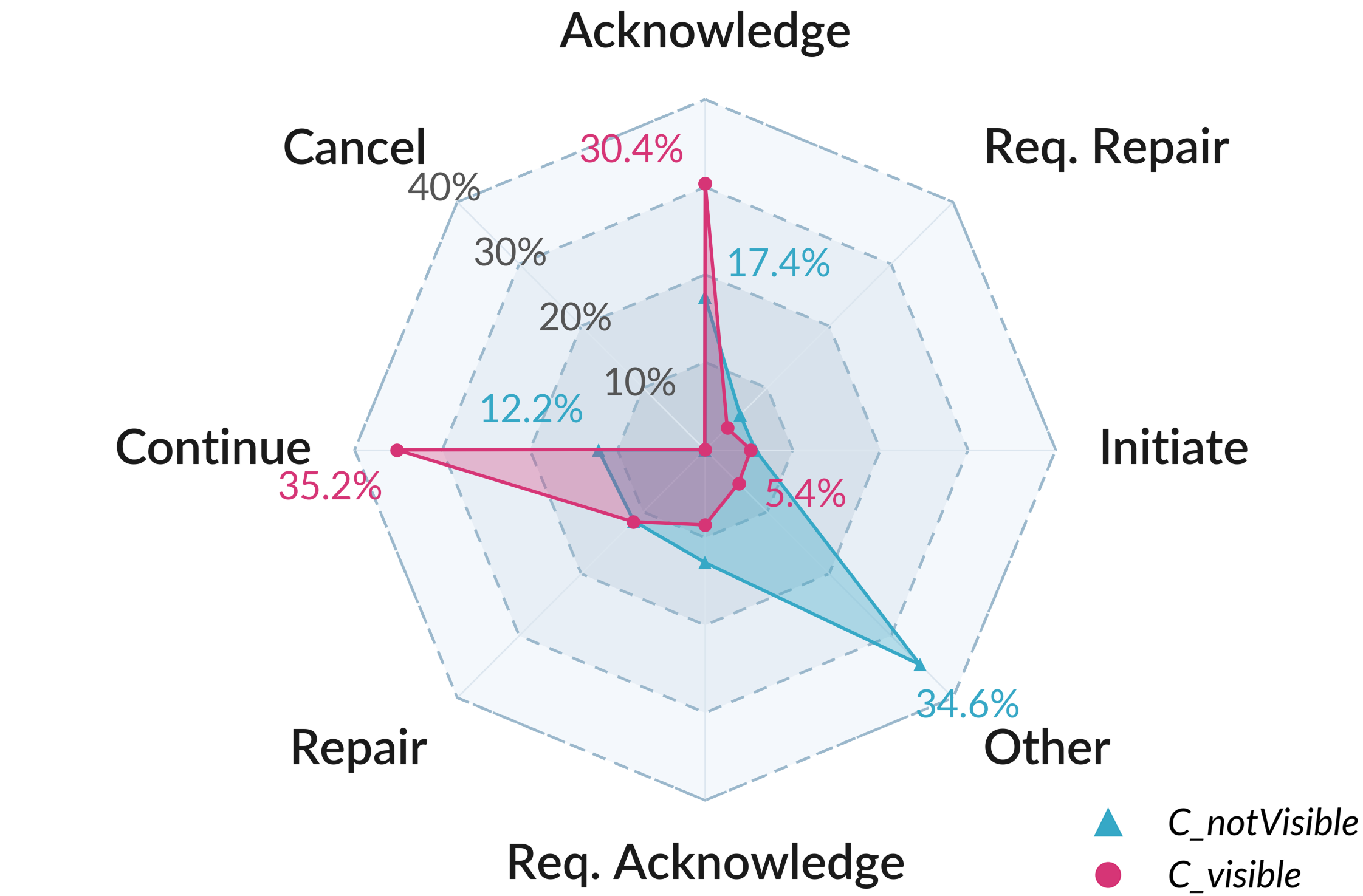}
    \caption{The relative proportion of each grounding act category under $C_{not\_visible}$ and $C_{visible}$.}
    \label{fig:grounding_dist}
\end{figure}

\begin{figure*}[htbp]
    \centering
    \includegraphics[width=1.00\linewidth]{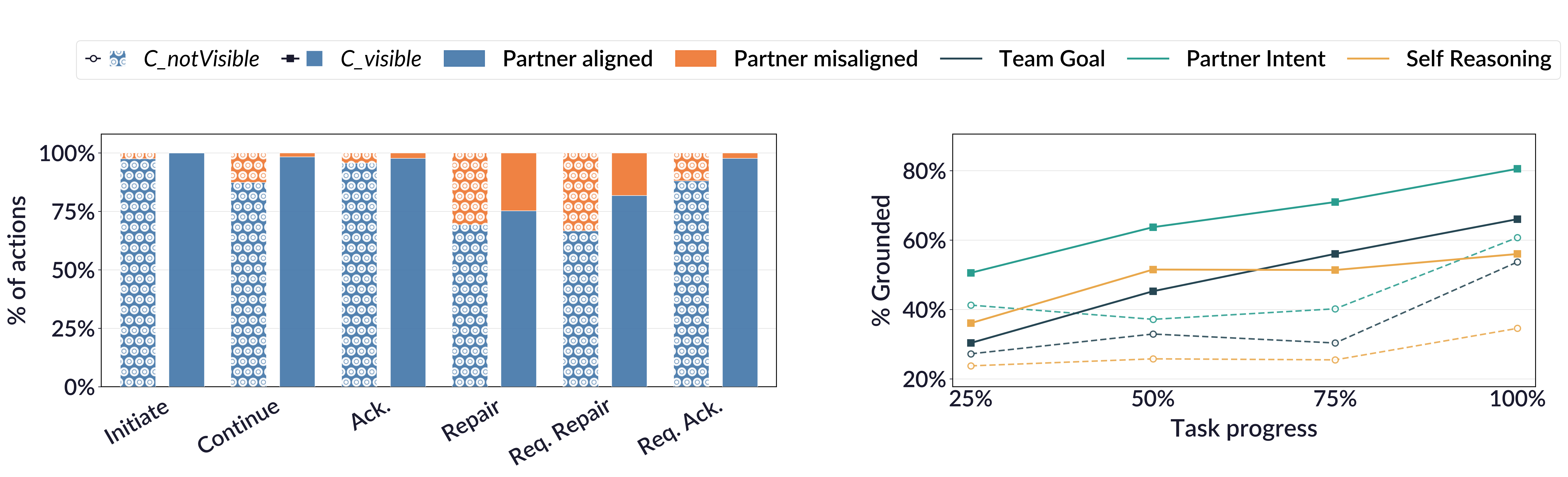}
    \caption{Relationships between grounding acts and mental model alignment. 
    Left: the proportion of perceived partner intent alignment within each grounding act category across $C_{not\_visible}$ and $C_{visible}$. 
    Right: the proportion of grounded team goal, partner intent, and self-reasoning annotations over task progress in the two conditions. 
}
    \label{fig:mm_charts}
\end{figure*}

\paragraph{Grounding Act Coding and Validation.}
To examine how participants establish, confirm, and repair shared understanding during collaboration, we additionally annotate \textbf{actions with grounding acts}. 
Three human annotators independently annotated 180 randomly sampled actions using the grounding act schema proposed by \citet{traum1995computational}, achieving an averaged inter-coder reliability of 0.81 measured by Fleiss’ $\kappa$ \cite{fleiss1973equivalence}.
We randomly partitioned the manually annotated actions into a few-shot set (8 actions) and a held-out validation set (172 actions). The few-shot set was used as in-context examples to prompt GPT-5.5 for automated annotation, while the validation set was reserved exclusively for evaluation.
On the held-out validation set, GPT-5.5 achieved a Fleiss’ $\kappa$ of 0.76 against human annotations, suggesting substantial agreement with human labels.

\paragraph{Behavioral and Mental Model Analysis.}
Figure~\ref{fig:grounding_dist} shows the distribution of grounding acts in $C_{not\_visible}$ and $C_{visible}$. 
In $C_{not\_visible}$, participants produced a higher proportion of \textit{Other} (non-grounding) actions. 
A review of the action logs suggests that drawing-related actions in $C_{not\_visible}$ were more likely to serve as individual exploration rather than mutually observable grounding acts.
By contrast, the higher proportions of \textit{Continue} and \textit{Acknowledge} acts in $C_{visible}$ suggest that canvas visibility helped participants maintain a clearer shared situation, allowing them to execute the grounded route plan with more explicit confirmation.

Figure~\ref{fig:mm_charts} further shows how mental model alignment relates to grounding acts and task progress. 
In the left bar chart, \textit{Acknowledge} acts are associated with higher perceived alignment, whereas repair-related acts, especially \textit{Repair} and \textit{Req. Repair}, are associated with higher perceived misalignment.
In the right line chart, alignment in team goal, partner intent, and self-reasoning generally increased as the task unfolded, with consistently higher alignment in $C_{visible}$ than in $C_{not\_visible}$. 
These trends suggest that participants' mental models became more aligned over time, and that access to the shared drawing state may have supported mutual understanding.
The systematic variation in mental model alignment across conditions and task stages provides a natural basis for evaluating whether LLMs can capture these collaborative dynamics in Sec.\ref{sec:benchmark}.

\section{Benchmark Experiment}
\label{sec:benchmark}

We evaluate how \datasetname can be leveraged to assess LLMs' ability to simulate humans' collaboration behavior and mental models through two complementary tasks:  \\ 
\indent \textbf{1. Next Behavior Prediction.} For a target participant, given the behavior trajectory history and persona profile, predict the next behavior. \\
\indent \textbf{2. Mental Model Prediction.} For a target participant, given the behavior trajectory history, mental model history, and persona profile, predict the participant's mental state in the next turn.



Next action prediction evaluates whether models can predict a human collaborator's next turn move from the preceding interaction context, reflecting their ability to simulate observable collaboration dynamics \cite{lu2025can, wang2025opera}.
Mental model prediction goes further by assessing whether models can infer the collaborator's next step underlying reasoning, which helps 
determine whether a model shows a genuine understanding of the 
collaboration state or merely fits surface-level trajectory patterns.

\subsection{Experiment Setup}

Our benchmark experiment includes two open-sourced models (Qwen3.6-35B-A3B and Llama 3.3 70B), two proprietary models (GPT 5.5 and Claude 4.6 Sonnet), and two finetuned models (Qwen3-4B and Qwen3-30B-A3B) on \datasetname. 
We evaluate these models under two approaches:
\paragraph{Persona-Based LLM.} In this setting, we evaluate whether general-purpose LLMs can simulate human collaborative behaviors when provided with only the participant's profile. 
Each model is given the participant persona, including demographic information and collaboration profiles, together with the interaction history up to the current action. 
We apply this setting to Qwen3.6-35B-A3B, Llama 3.3 70B, GPT-5.5, and Claude 4.6 Sonnet. 
The full prompts are provided in Appendix~\ref{app:prompts}.
\paragraph{Fine-Tuned LLM.} In this setting, we examine whether smaller models can benefit from supervision on \datasetname. 
We fine-tune Qwen3-4B and Qwen3-30B-A3B on the training split and evaluate their performance on the same next action prediction and mental model prediction tasks. 
Hyper-parameters are reported in Appendix \ref{app:hyper}

For the next action prediction task, we compare three settings to explore the effectiveness of human-annotated mental models. 
In the default setting, the model predicts the next action from only the interaction history. 
In the second, the model is given the participant's annotated mental model before the target action, denoted as 
\textit{\{+Mental Model\}}.
Including ground-truth mental model annotations evaluates whether explicit human-annotated mental model information can improve model performance.
We also consider a Chain of Thought setting, denoted as \textit{\{+CoT\}}, in which the model first generates a rationale from the interaction history and then predicts the next action conditioned on that rationale. 
Finally, we experiment a two-stage ReAct-inspired baseline, denoted as \textit{{+Pred. MM}}, in which the model first predicts humans' mental model from the observed interaction history and then predicts the next action conditioned on the predicted mental model.
The full results are shown in Appendix \ref{app:experiments}. 

\begin{table*}[th!]
\centering
\small
\resizebox{1.00\linewidth}{!}{
\begin{tabular}{l|llll|llll|llll|llll}
\toprule
\multirow{3}{*}{\centering\textbf{Model}}
& \multicolumn{4}{c|}{$\boldsymbol{Accuracy_{\text{Action\_Type}}}$}
& \multicolumn{4}{c|}{$\boldsymbol{Recall_{\text{Action\_Type}}}$}
& \multicolumn{4}{c|}{$\boldsymbol{SBERT_{\text{Message}}}$}
& \multicolumn{2}{c}{$\boldsymbol{Accuracy_{\text{Drawing}}}$} \\
& \multicolumn{2}{c}{$C_{visible}$}
& \multicolumn{2}{c|}{$C_{not\_visible}$}
& \multicolumn{2}{c}{$C_{visible}$}
& \multicolumn{2}{c|}{$C_{not\_visible}$}
& \multicolumn{2}{c}{$C_{visible}$}
& \multicolumn{2}{c|}{$C_{not\_visible}$}
& \multicolumn{1}{c}{$C_{visible}$}
& \multicolumn{1}{c}{$C_{not\_visible}$} \\
& G & F & G & F
& G & F & G & F
& G & F & G & F
& F & F \\
\midrule
Qwen3-35B-A3B          & 1.00 & 0.46 & 1.00 & 0.54 & 1.00 & 0.34 & 1.00 & 0.30 & 0.22 & 0.25 & 0.28 & 0.27 & 0.43 & 0.45 \\
~~~~~\textit{+Mental Model}   & 1.00 & 0.48\up & 1.00 & 0.55\up & 1.00 & 0.38\up & 1.00\up & 0.31\up & 0.21 & 0.27\up & 0.28 & 0.29\up & 0.44\up & 0.43 \\
\midrule[0.4pt]
Llama 3.3 70B          & 1.00 & 0.44 & 1.00 & 0.51 & 1.00 & 0.32 & 1.00 & 0.28 & 0.22 & 0.23 & 0.31 & 0.27 & 0.57 & 0.45 \\
~~~~~\textit{+Mental Model}   & 1.00 & 0.51\up & 1.00 & 0.51 & 1.00 & 0.28 & 1.00 & 0.30\up & 0.22 & 0.25\up & 0.30 & 0.29\up & 0.46 & \textbf{0.56}\up \\
\midrule[0.4pt]
GPT-5.5                & 1.00 & 0.56 & 1.00 & 0.59 & 1.00 & 0.44 & 1.00 & 0.35 & 0.24 & 0.32 & 0.33 & 0.36 & \textbf{0.55} & 0.43 \\
~~~~~\textit{+Mental Model}   & 1.00 & \textbf{0.58}\up & 1.00 & \textbf{0.61}\up & 1.00 & \textbf{0.46}\up & 1.00 & \textbf{0.37}\up & \textbf{0.25}\up & 0.17 & 0.33 & \textbf{0.38}\up & \textbf{0.55} & 0.47\up \\
\midrule[0.4pt]
Claude 4.6 Sonnet      & 1.00 & 0.47 & 1.00 & 0.54 & 1.00 & 0.36 & 1.00 & 0.30  & 0.23 & 0.29 & 0.31 & 0.31 & 0.45 & 0.44 \\
~~~~~\textit{+Mental Model}   & 1.00 & 0.51\up & 1.00 & 0.55\up & 1.00 & 0.39\up & 1.00 & 0.31\up & 0.23 & 0.31\up & 0.31 & 0.33\up & 0.43 & 0.53\up \\
\midrule[0.4pt]
Qwen3-4B FT  & 1.00 & 0.56 & 1.00 & 0.54 & 1.00 & 0.37 & 1.00 & 0.31 & 0.21 & 0.35 & 0.23 & 0.23 & 0.47 & 0.44 \\
Qwen3-30B-A3B FT & 1.00 & 0.52 & 1.00 & 0.52 & 1.00 & 0.30 & 1.00 & 0.27 & 0.20 & \textbf{0.37} & 0.22 & 0.26 & 0.54 & 0.55 \\
\bottomrule
\end{tabular}}
\caption{Action Type Accuracy, Action Type Recall, Message SBERT, and Follower's Drawing Accuracy across six models in Guide (G) and Follower (F) roles under $C_{visible}$ and $C_{not\_visible}$. \textbf{Bolded} numbers (excluding Guide's Action Type Accuracy and Recall) indicate the best performance for each role and canvas visibility condition.}
\label{tab:next_action}
\end{table*}

\subsection{Evaluation}
\label{sec:eval}
We evaluate model outputs at two levels of granularity. 
At the category level, we report \textbf{accuracy} and \textbf{recall} for the predicted next action type and mental model category. 
At the content level, we evaluate semantic similarity using \textbf{SBERT} \cite{reimers2019sentence}. For message actions, we compare the generated message to the ground-truth message; for mental model prediction, we compare the generated rationale to the participant's action-level annotation. 
Full results including \textbf{ROUGE-L} \cite{lin2004rouge} are presented in Appendix \ref{app:experiments}.

To evaluate models' drawing trace accuracy, we use a distance-weighted score. 
Each predicted ink cell is scored by its Chebyshev distance to the ground-truth route: 1 if on the route, $2/3$ at distance 1, $1/3$ at distance 2, and 0 otherwise. 
We report the average score over all predicted ink cells.

\begin{table*}[t]
\centering
\small
\resizebox{1.00\linewidth}{!}{
\begin{tabular}{l|cccc|cccc|cccc|cccc}
\toprule
\multirow{3}{*}{\centering\textbf{Model}}
& \multicolumn{4}{c|}{$\boldsymbol{Accuracy_{\text{Team\_Goal}}}$}
& \multicolumn{4}{c|}{$\boldsymbol{Accuracy_{\text{Partner\_Intent}}}$}
& \multicolumn{4}{c|}{$\boldsymbol{Accuracy_{\text{Self\_Reasoning}}}$}
& \multicolumn{4}{c}{$\boldsymbol{SBERT_{\text{Rationale}}}$} \\
& \multicolumn{2}{c}{$C_{visible}$} & \multicolumn{2}{c|}{$C_{not\_visible}$}
& \multicolumn{2}{c}{$C_{visible}$} & \multicolumn{2}{c|}{$C_{not\_visible}$}
& \multicolumn{2}{c}{$C_{visible}$} & \multicolumn{2}{c|}{$C_{not\_visible}$}
& \multicolumn{2}{c}{$C_{visible}$} & \multicolumn{2}{c}{$C_{not\_visible}$} \\
& G & F & G & F
& G & F & G & F
& G & F & G & F
& G & F & G & F \\
\midrule
Qwen3-35B-A3B            & \textbf{0.48} & 0.69 & 0.41 & 0.64 & \textbf{0.58} & 0.73 & 0.38 & 0.67 & \textbf{0.31} & 0.56 & 0.32 & 0.52 & \textbf{0.41} & 0.46 & 0.44 & 0.52 \\
Llama 3.3 70B            & 0.43 & 0.71 & \textbf{0.56} & 0.64 & 0.38 & 0.76 & 0.34 & 0.72 & 0.23 & 0.55 & 0.31 & 0.53 & \textbf{0.41} & 0.47 & 0.44 & 0.54 \\
GPT-5.5                  & 0.41 & 0.72 & 0.35 & 0.68 & 0.37 & 0.75 & 0.36 & 0.70 & 0.29 & 0.60 & 0.32 & 0.59 & 0.40 & 0.51 & \textbf{0.45} & 0.55 \\
Claude 4.6 Sonnet        & \textbf{0.48} & 0.75 & 0.45 & 0.68 & 0.51 & 0.76  & 0.45 & 0.71 & 0.27 & 0.56 & 0.30 & 0.55 & \textbf{0.41} & 0.52 & \textbf{0.45} & 0.55 \\
\midrule
Qwen3-4B FT      & 0.37 & \textbf{0.81} & 0.51 & \textbf{0.88} & 0.40 & \textbf{0.84} & \textbf{0.47} & \textbf{0.84} & 0.28 & \textbf{0.65} & 0.30 & \textbf{0.70} & 0.37 & \textbf{0.76} & 0.33 & 0.64 \\
Qwen3-30B-A3B FT & 0.47& 0.55 & 0.39 & 0.55 & 0.38 & 0.78 & 0.46 & 0.77 & 0.29 & 0.54 & \textbf{0.33} & 0.54 & 0.34 & 0.61 & 0.38 & \textbf{0.66} \\
\bottomrule
\end{tabular}}
\caption{Team Goal Accuracy, Partner Intent Accuracy, Self-Reasoning Accuracy, and Rationale SBERT across six models in Guide (G) and Follower (F) roles under $C_{visible}$ and $C_{not\_visible}$. \textbf{Bolded} numbers indicate the best performance for each role and canvas visibility condition.}
\label{tab:mental_model}
\end{table*}

\subsection{Result and Analysis}

\subsubsection{Next Action Prediction}

Table \ref{tab:next_action} shows model performance on the next action prediction task. 
Across all models, Guide action type prediction is perfect, as expected, since the Guide's action space is limited to \texttt{message}. 
Follower action prediction, by contrast, is substantially harder, which is consistent with Followers alternating between interpreting messages, drawing, correcting, and grounding their understanding.

Models generally perform better in $C_{not\_visible}$ than in $C_{visible}$ across action type, message, and drawing prediction. 
For example, GPT-5.5 achieves higher Follower $Accuracy_{\text{Action\_Type}}$ in $C_{not\_visible}$, and its Follower's $SBERT_{\text{Message}}$ rises from 0.17 to 0.38 with mental model input. 
For drawing, GPT-5.5 reaches around 0.55 in $C_{visible}$, while in $C_{not\_visible}$, mental model input improves $Accuracy_{\text{Drawing}}$ from 0.43 to 0.47. 
The greater behavioral variability in $C_{visible}$ may explain this pattern, since Guides who can observe the Follower's live canvas tend to rely more on visually grounded corrections, interruptions, and fine-grained coordination, making the next action harder to infer from textual history alone.

Adding mental model input generally improves model performance in Followers' action prediction. 
For GPT-5.5, mental model input slightly improves Follower action type accuracy and recall in both conditions, and improves drawing accuracy in $C_{not\_visible}$. 
This result suggests that Followers' behavior is more grounded in latent reasoning states (e.g., interpreting instructions) than Guides', whose actions are more anchored to the visible canvas and task goal.
Notably, smaller models fine-tuned on \datasetname come close to large proprietary models, indicating that targeted supervision on \datasetname can effectively close the gap with larger models.

\subsubsection{Mental Model Prediction}

Table \ref{tab:mental_model} shows model performance on the mental model prediction task. Follower mental models are easier to predict than Guides' across all three dimensions. For example, Claude 4.6 Sonnet achieves 0.75 $Accuracy_{\text{Team\_Goal}}$ for the Follower under $C_{visible}$, but only 0.48 for the Guide, with similar gaps for partner intent and self-reasoning. 
Although the Guide's action space is limited to \texttt{message}, the Guide's underlying reasoning likely involves richer spatial planning and partner monitoring that are hard to infer from interaction history alone, which may account for the asymmetry.

Across roles and conditions, self-reasoning is the hardest dimension to predict, whereas team goal and partner intent are more predictable. Team goal and partner intent are often reflected in shared task progress and dialogue content, while self-reasoning captures participant-specific motivations that may not be explicitly expressed. As a result, current LLMs appear better at approximating shared components of mental model awareness than inferring private reasoning that varies across participants.

Notably, fine-tuned Qwen-3-4B achieves the strongest performance in Followers’ mental model prediction and $SBERT_{\text{Rationale}}$, whereas prompt-based models' Follower $SBERT_{\text{Rationale}}$ scores stay within a narrow range of 0.40--0.55.
This pattern suggests that \datasetname’s mental model annotations provide useful supervision for learning collaboration-relevant reasoning when participant states are reflected in the interaction history. 
Overall, no single model consistently performs best across all conditions, and the low self-reasoning accuracy highlights private mental model inference as a central challenge in \datasetname.

\section{Discussion}

Our results show that mental model annotations provide useful signals for modeling collaborative behavior. 
In next action prediction, adding these annotations improves some models' performance, but the gains are inconsistent, indicating that mental models encode signals that current LLMs do not reliably leverage. 
The mental model prediction results reinforce this interpretation, given that the shared components (e.g., team goal and partner intent) are easier to infer than self-reasoning, suggesting that models handle publicly grounded collaboration states better than private reasoning.

The two experiments suggest a role-specific dissociation between behavior prediction and mental model prediction. 
Guide mental models are harder to infer because they involve less observable reasoning about route planning and Follower progress.
Followers show the opposite pattern: their broader action space makes behavior prediction harder, but their mental models are more directly shaped by the Guide’s explicit instructions. 
This result suggests that observable behavior and mental models provide complementary signals, so success on one does not necessarily imply success on the other.

Importantly, next action prediction is not the end goal of \datasetname; it serves as a diagnostic for whether models can simulate the observable layer of collaboration. 
The deeper challenge lies in building agents that maintain accurate mental models throughout the collaboration.
\datasetname provides a foundation for developing such agents by supplying the process-level supervision signals that current training paradigms lack.

\section{Conclusion}

In this work, we present \datasetname, an authentic human collaboration dataset that captures both human collaboration behaviors and the underlying action-level mental models,
including how humans reason about their team goals, partners’ intentions, and self-reasoning over time.
We demonstrate the utility of \datasetname through next action prediction and mental model prediction. Our results show that mental model annotations provide signals beyond interaction history alone, and that shared mental model components are substantially easier to predict than private self-reasoning.
This observation highlights a fundamental gap in models' ability to infer the cognitive processes that drive effective collaboration.
By grounding agent evaluation in authentic human collaboration data with theory-informed mental model annotations, \datasetname opens a pathway toward developing LLM agents that can serve as genuine collaborative partners rather than sophisticated task-solvers.

\section*{Acknowledgment}
This work was supported in part by a Microsoft Research Agentic AI Research and Innovation Award. Any opinions, findings, and conclusions or recommendations expressed in this material are those of the authors and do not necessarily reflect the views of the sponsors.

\section{Limitations}
This work has several limitations, which we discuss alongside the design choices that mitigate them.

First, our annotation framework relies in part on post-session retrospective reports, which are susceptible to recall bias and post-hoc rationalization. 
We mitigate this concern through two design choices: the \textit{in-session} checkpoints capture real-time mental states at three task stages and serve as memory anchors during \textit{post-session} annotation, and the annotation interface presents action-level screenshots alongside the interaction trajectory to support context-specific recall. 
Future work could explore concurrent think-aloud protocols or physiological measures to further validate the fidelity of retrospective annotations.

Second, the \datasetname dataset comprises 25 sessions from 50 participants, which is modest compared to some large-scale NLP benchmarks. 
Nevertheless, the detailed action-level annotation of theory-grounded mental models and rationales partially compensates for the session count, which yields 2,987 individually annotated data points with both structured labels and free-form rationales. 
In addition, our participants include both native and non-native English speakers, and we do not control for proficiency level in the current analysis. 

Third, \datasetname is built from the Map Task, a single controlled task domain selected for its theoretical grounding in social science research and its natural combination of language and workspace actions~\cite{anderson1991hcrc}. 
While the controlled setting allows us to isolate collaboration variables, real-world collaboration often involves longer time horizons and more complex interaction constraints and social dynamics. 
Extending the annotation framework to other collaborative tasks, such as collaborative writing, programming, or decision-making, would strengthen claims about the generalizability of both the dataset and the benchmark findings. 
A promising direction for future work is to study how models can adaptively construct and update mental models in diverse, real-world, domain-specific environments.

Fourth, our benchmark evaluates six LLMs under persona-based prompting and supervised fine-tuning on \datasetname, but does not include models fine-tuned on other collaborative dialogue datasets (e.g., CaSiNo~\cite{chawla-etal-2021-casino}, DealNoDeal~\cite{lewis-etal-2017-deal}) or models trained with alternative alignment approaches such as reinforcement learning from human feedback. 
Including such baselines would help disentangle whether performance gaps stem from the absence of collaboration-specific training signals or from architectural limitations of current models. 
In addition, current language models remain limited in interpreting drawing actions and map states. 
Although we represent maps and drawing trajectories in structured text-based formats, these representations may not fully capture the spatial relationships that human participants perceive visually. 
Future work could explore multimodal models that jointly process visual and textual input to better represent the spatial task state.


\bibliography{custom,yuxuan}

\clearpage
\appendix



\section{Mental Model Annotation}

\subsection{Annotation Schema}
\label{app:schema}
Table \ref{tab:annotation-schema} presents the annotation schema we used to collect \datasetname, along with the collaboration theories that inform its design.

\begin{table*}[t]
\centering
\small
\renewcommand{\arraystretch}{1.25}
\begin{tabularx}{\textwidth}{@{}p{0.03\textwidth} p{0.10\textwidth} p{0.16\textwidth} X p{0.21\textwidth}@{}}
\toprule
\textbf{ID} & \textbf{Category} & \textbf{Question} & \textbf{Response Options} & \textbf{Theoretical Grounding} \\
\midrule
Q1 & Team Goal &
At this moment, my partner and I were: &
\begin{tabular}[t]{@{}p{0.01\textwidth}p{0.35\textwidth}@{}}
t1 & Still figuring out what we needed to do \\
t2 & Working toward a shared understanding \\
t3 & Clear on what to do and working on it \\
t4 & Something was unclear and we were working it out \\
-- & Other \\
\end{tabular} &
\textit{Shared Mental Models} \cite{cannon1993shared}; team task awareness in human--AI teaming \cite{andrews2023role}. \\
\addlinespace[2pt]
Q2 & Partner Intent &
At this moment, I thought my partner: &
\begin{tabular}[t]{@{}p{0.01\textwidth}p{0.39\textwidth}@{}}
p1 & Understood the situation; on the same page \\
p2 & Probably understood, but I was not fully sure \\
p3 & Is waiting for more information \\
p4 & Misunderstood; not aligned \\
p5 & Gave no clear signal either way \\
-- & Other \\
\end{tabular} &
\textit{Shared Mental Models} \cite{cannon1993shared}; \textit{Partner Models} and theory of mind in dialogue \citep{clark1991grounding, doyle2019mapping}. \\
\addlinespace[2pt]
Q3 & self-reasoning &
At this moment, my action was driven by: &
\begin{tabular}[t]{@{}p{0.01\textwidth}p{0.39\textwidth}@{}}
r1 & Executing an agreed-upon plan \\
r2 & Exploring on my own to gather information \\
r3 & Confirming the situation with my partner \\
r4 & Grounding -- sharing/requesting info to align \\
r5 & Repairing a mistake or misunderstanding \\
r6 & Waiting for more information \\
-- & Other \\
\end{tabular} &
\textit{Grounding acts} and conversational grounding \citep{clark1991grounding,traum1995computational}; self-component of \textit{Shared Mental Models} and metacognitive action selection in joint activity \citep{klein2005common}. \\
\addlinespace[2pt]
Q4 & Alignment &
At this moment, my partner and I were on the same page (Yes / No). If No, why: &
\begin{tabular}[t]{@{}p{0.01\textwidth}p{0.39\textwidth}@{}}
n1 & Different understanding of the task goal \\
n2 & Different understanding of the current state \\
n3 & One of us was missing key information \\
n4 & Communication was unclear or ambiguous \\
n5 & Technical or interface issue got in the way \\
\end{tabular} &
\textit{Common ground} \citep{clark1991grounding}; breakdowns in common ground for joint activity \citep{klein2005common}; coordination breakdowns \citep{schmidt1992taking}. \\
\bottomrule
\end{tabularx}
\caption{Annotation schema used to collect \datasetname.}
\label{tab:annotation-schema}
\end{table*}

\subsection{Annotation Validation}
\label{app:mm_annotation_validation}

To examine whether the retrospective annotations retained participant-specific information from the collaboration's corresponding moments, we compared their in-session checkpoint responses with their post-session annotated mental-model rationales.
For each checkpoint, we computed its mean SBERT score across all post-session-annotated rationales from the same participant within the corresponding interval (\textsc{Self}). 
As a comparison, we computed its similarity with rationales from other participants at the corresponding task position (\textsc{Others}).
Table~\ref{tab:mm_annotation_validation} summarizes the results.

Across all conditions, in-session checkpoint responses were more similar to participants' retrospective annotations than to those of other participants.
The mean similarity difference was approximately 0.10 overall and remained consistent across the two experimental conditions.
These differences were statistically significant ($p=0.0005$), suggesting that the retrospective annotations preserve participant-specific information reflected in participants' mental models during the collaboration.

\begin{table}
    \centering
    \small
    \begin{tabular}{lrrrr}
        \toprule
        \textbf{Condition} 
        & \textbf{$N$}
        & \textbf{SBERT$_{\text{Self}}$}
        & \textbf{SBERT$_{\text{Others}}$}
        & \textbf{$p$} \\
        \midrule
        Overall                  & 139 & 0.43 & 0.33 & 0.000$^{*}$ \\
        $C_{\text{not-visible}}$ & 69  & 0.42 & 0.32  & 0.000$^{*}$ \\
        $C_{\text{visible}}$     & 70  & 0.43 & 0.33  & 0.000$^{*}$ \\
        \bottomrule
    \end{tabular}
    \caption{
    Alignment between participants' in-session checkpoint responses and post-session mental-model annotations.
    \textsc{Self} denotes the mean SBERT cosine similarity with the participant's own annotations, while \textsc{Others} denotes similarity with annotations from other participants at the corresponding task position.
    }
\label{tab:mm_annotation_validation}
\end{table}

\section{Questionnaire}
\label{app: survey}

\subsection{Demographic Information Questionnaire Items and Results}

Table~\ref{tab:demographic_questions} lists the demographic question items used in our study.
Our study included 66 participants. 
Participants were relatively balanced in gender (32 female, 31 male, 2 non-binary/third gender, and 1 preferred not to disclose). 
Most participants were between 18–34 years old (28 aged 18–24; 33 aged 25–34), with 4 aged 35–44 and 1 aged 45–54. 
34 participants held a master's degree, 24 a bachelor's degree, 4 a high school diploma or equivalent, 2 a doctoral degree, 1 an associate degree, and 1 had less than a high school education.

\begin{table}[h]
\centering
\small
\begin{tabular}{p{0.4\linewidth} p{0.48\linewidth}}
\toprule
\textbf{Question} & \textbf{Options} \\
\midrule
What is your gender? 
& Male; Female; Non-binary / third gender; Prefer not to say \\
\midrule
What is your age group? 
& 18--24; 25--34; 35--44; 45--54; 55+ \\
\midrule
What is the highest level of education you have earned? 
& Less than high school; High school or equivalent; Associate degree; Bachelor's degree; Master's degree; Doctoral degree \\
\bottomrule
\end{tabular}
\caption{Demographic questions used in the pre-study questionnaire.}
\label{tab:demographic_questions}
\end{table}

\subsection{Collaboration Style Questionnaire Items}

We use items from TeamQ \cite{britton2017assessing} to collect participants' perceived collaboration styles.
They responded to each item using a 5-point frequency scale. Table \ref{tab:teamwork_behavior_questionnaire} shows the items used in our pre-study survey.

\begin{table*}[t!]
\centering
\small
\begin{tabular}{p{0.2\linewidth} p{0.73\linewidth}}
\toprule
\textbf{Construct} & \textbf{Item} \\
\midrule
Task Contribution 
& Participate actively and accept a fair share of the group work. \\

Task Contribution 
& Work skillfully on assigned tasks and complete them on time. \\

Feedback 
& Give timely, constructive feedback to team members in the appropriate format. \\

Communication 
& Communicate actively and constructively. \\

Inclusiveness 
& Encourage all perspectives to be considered and acknowledge contributions of others. \\

Integration 
& Constructively build on contributions of others and integrate own work with work of others. \\

Coordination 
& Take on an appropriate role in the group, e.g., leader or note taker. \\

Coordination 
& Clarify goals and plan the project. \\

Coordination 
& Report to team on progress. \\

Interpersonal Expression 
& Ensure consistency between words, tone, facial expression, and body language. \\

Team Climate 
& Express positivity and optimism about team members and project. \\

Conflict Management 
& Display appropriate assertiveness: neither dominating, submissive, nor passive aggressive. \\

Conflict Management 
& Contribute appropriately to healthy debate. \\

Conflict Management 
& Respond to and manage direct/indirect conflict constructively and effectively. \\
\bottomrule
\end{tabular}
\caption{Collaboration tendency questionnaire items used in the pre-study questionnaire.}
\label{tab:teamwork_behavior_questionnaire}
\end{table*}

\section{Map Material for Data Collection}
\label{app:map}

\begin{figure}
    \centering
    \includegraphics[width=0.9\linewidth]{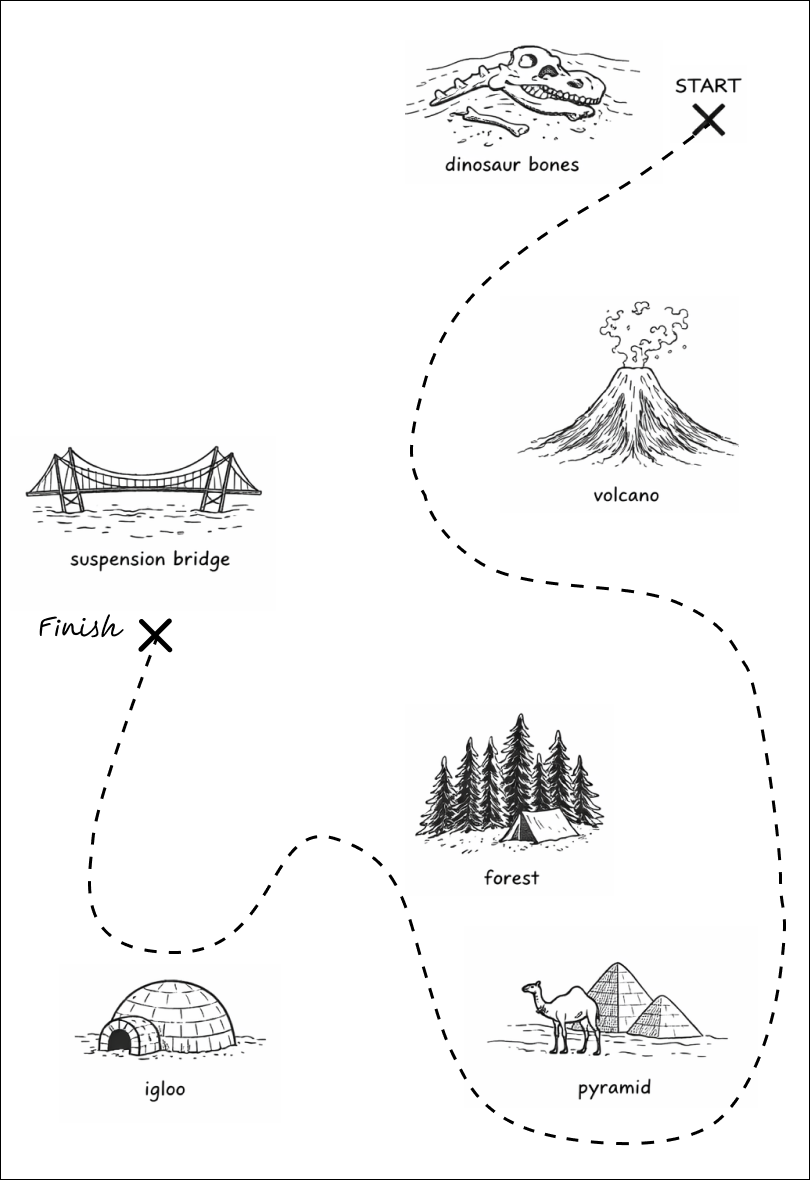}
    \caption{Guide map used in the Map Task. The guide has access to the target route and provides instructions to help the follower reproduce the route.}
    \label{fig:guide_map}
\end{figure}

\begin{figure}
    \centering
    \includegraphics[width=0.9\linewidth]{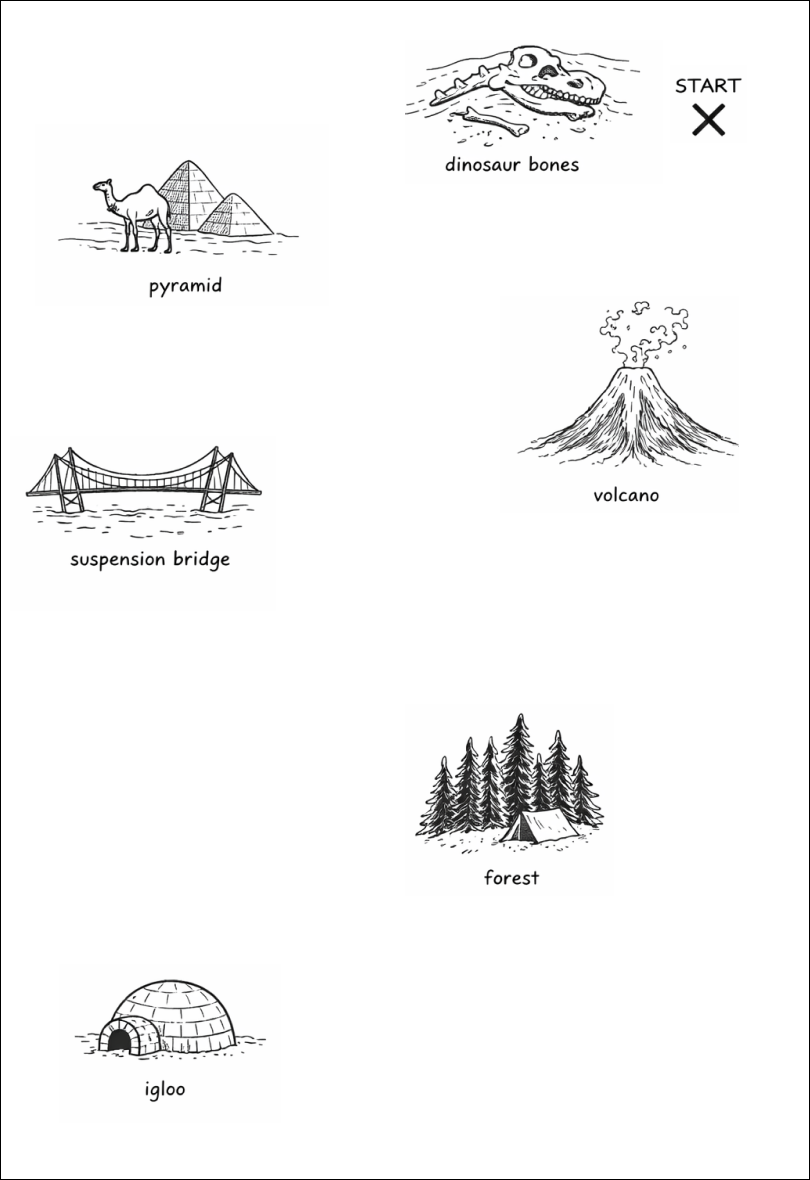}
    \caption{Follower map used in the Map Task. The follower sees the map landmarks but does not have access to the target route.}
    \label{fig:follower_map}
\end{figure}

\begin{figure}
    \centering
    \includegraphics[width=0.9\linewidth]{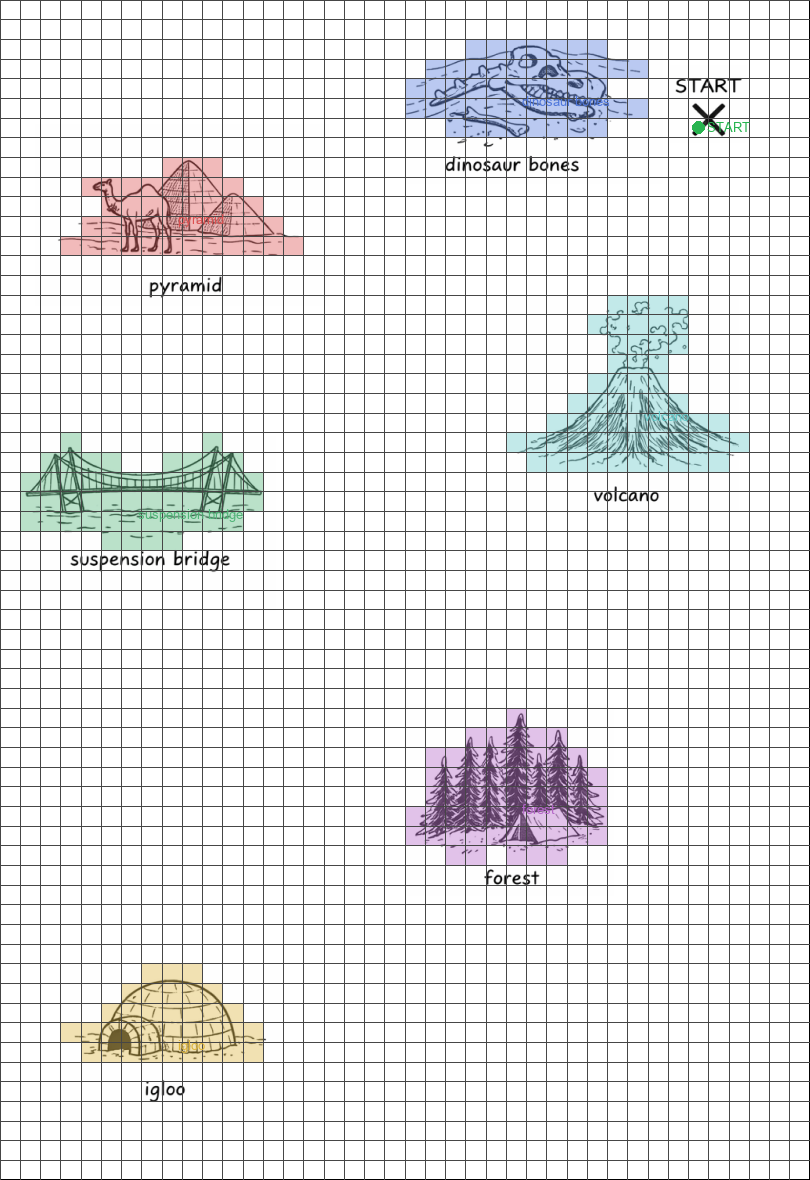}
    \caption{Grid-based map representation used for LLM-readable standardization. Landmark positions are manually annotated and converted into structured spatial representations.}
    \label{fig:grid_map}
\end{figure}

We adapted the map materials from \citet{anderson1991hcrc}. 
In the pilot studies, we used the original maps from their work. However, because the original Map Task was conducted through in-person interaction, transferring the task to a computer-mediated setting increased task difficulty and resulted in longer completion times. 
To better suit our study context, we retained the original map style but reduced the number of landmarks and simplified the route structure. Figures~\ref{fig:guide_map} and~\ref{fig:follower_map} present the map materials used in our data collection.
To standardize the map images for LLM comprehension, we convert each map into a grid-based representation and manually annotate the positions of all landmarks (Figure~\ref{fig:grid_map}). 
We then encode the map content in a structured JSON format, which provides the LLM with explicit spatial information about the grid, start point, and landmark regions:

\begin{verbatim}
{
  "grid_size": [...],
  "start_cell": [...],
  "landmarks": {
    "pyramid": {
      "summary": {
        "bbox": {
          "row_min": ...,
          "row_max": ...,
          "col_min": ...,
          "col_max": ...
        },
        "corners": {
          "top_left": [...],
          "top_right": [...],
          "bottom_left": [...],
          "bottom_right": [...]
        },
        "centroid": [...],
        "boundary_cells": [...]
      },
      "cells": [...],
      "type": "blocked"
    },
    "suspension bridge": {
      ...
    },
    ...
  }
}
\end{verbatim}

\section{Dataset Partition and Distribution}
\label{app:distribution}
We split \datasetname at the session level to avoid data leakage across train and test splits.
Because the two condition settings change participants' available evidence, we construct splits separately for $C_{not\_visible}$ and $C\_visible$. 
We use an approximately 4:1 train/test split within each condition.
For $C_{not\_visible}$, we assign 9 sessions to training and 3 sessions to test. 
For $C\_visible$, we assign 10 sessions to training and 3 sessions to test, resulting in 19 training sessions and 6 test sessions overall.

To make the training and test sets comparable, we select test sessions through distribution matching rather than random sampling. 
For each session, we compute role-specific proportions over action types (\textit{draw}, \textit{erase}, \textit{message}, \textit{reset}, and \textit{undo}) and over mental-model labels along three dimensions: team goal, partner intent, and self reasoning. 
We concatenate the Guide and Follower proportion vectors into a shared session representation, with unseen labels assigned a proportion of zero. 
Within each condition, we enumerate all candidate subsets of three test sessions and choose the subset that minimizes the following objective:
\[
\mathcal{L}
=
\sum_d \left|\mu^{\mathrm{train}}_d - \mu^{\mathrm{test}}_d\right|
+
\sum_d \left|\sigma^{\mathrm{train}}_d - \sigma^{\mathrm{test}}_d\right|,
\]
where $d$ indexes each feature dimension, and $\mu_d$ and $\sigma_d$ denote the across-session mean and standard deviation of session-level proportions. 
Because the objective compares both central tendency and cross-session variability, the resulting split preserves the overall composition of action types and mental-model labels across training and test sets.

Table~\ref{tab:stats} reports both the action type and the resulting mental-model distribution. 
The train and test sets preserve similar label composition across the three dimensions, while retaining lower-frequency labels related to uncertainty, repair, and waiting for information.

\section{Hyper-Parameters and Experiment
Settings}
\label{app:hyper}

All eight jobs share the same training configuration. We train with a sequence length of 65,536, a global batch size of 32, and a micro-batch size of 1, for 3 epochs in total. We use the Distributed Fused Adam optimizer with adam\_beta2 = 0.95, a peak learning rate of 5e-5 (min-lr 0.0) under a cosine annealing schedule with no warmup iterations, and bf16 mixed precision. Checkpoints are saved once per epoch (save-interval = 1), with save-optim disabled. The random seed is fixed to 5678. For training efficiency, we enable sequence parallelism, the distributed optimizer, as well as gradient-reduce and parameter-gather overlap.

We use different parallelism configurations for the two model scales. For Qwen-3-4B, we adopt tensor parallelism (TP) of 4, context parallelism (CP) of 2, and pipeline parallelism (PP) of 1, running on 8 GPUs (1 node). For  Qwen3-30B-A3B, we use TP = 4, CP = 2, PP = 4, and expert parallelism (EP) of 2, running on 32 GPUs across 4 nodes.

\section{Complete Experiment Results}
\label{app:experiments}

Table \ref{tab:next_action_full} reports the complete experimental results for the next action prediction task, additionally including Rouge-L scores and results under Chain-of-Thought prompting. Table \ref{tab:mental_model_full} reports the complete experimental results for the mental model prediction task, additionally including Rouge-L scores.

\begin{table*}[t]
\centering
\small
\resizebox{1.00\linewidth}{!}{
\begin{tabular}{l|cc|cc|cccc|cccc|cc}
\toprule
\multirow{3}{*}{\centering\textbf{Model}}
& \multicolumn{2}{c|}{$\boldsymbol{Accuracy_{\text{Action\_Type}}}$}
& \multicolumn{2}{c|}{$\boldsymbol{Recall_{\text{Action\_Type}}}$}
& \multicolumn{4}{c|}{$\boldsymbol{SBERT_{\text{Message}}}$}
& \multicolumn{4}{c|}{$\boldsymbol{ROUGE_{\text{Message}}}$}
& \multicolumn{2}{c}{$\boldsymbol{Accuracy_{\text{Drawing}}}$} \\

& \multicolumn{1}{c}{$C_{visible}$}
& \multicolumn{1}{c|}{$C_{not\_visible}$}
& \multicolumn{1}{c}{$C_{visible}$}
& \multicolumn{1}{c|}{$C_{not\_visible}$}
& \multicolumn{2}{c}{$C_{visible}$}
& \multicolumn{2}{c|}{$C_{not\_visible}$}
& \multicolumn{2}{c}{$C_{visible}$}
& \multicolumn{2}{c|}{$C_{not\_visible}$}
& \multicolumn{1}{c}{$C_{visible}$}
& \multicolumn{1}{c}{$C_{not\_visible}$} \\

& F & F
& F & F
& G & F & G & F
& G & F & G & F
& F & F \\
\midrule

Qwen3-35B-A3B
& 0.46 & 0.54
& 0.34 & 0.30
& 0.22 & 0.25 & 0.28 & 0.27
& 0.07 & 0.10 & 0.09 & 0.12
& 0.43 & 0.45 \\

~~~~~\textit{+CoT}
& 0.51 & 0.53
& 0.39 & 0.30
& 0.21 & 0.23 & 0.27 & 0.27
& 0.06 & 0.08 & 0.07 & 0.10
& 0.43 & 0.38 \\

~~~~~\textit{+Pred. MM}
& 0.50 & 0.53
& 0.27 & 0.23
& 0.21 & 0.25 & 0.26 & 0.26
& 0.07 & 0.10 & 0.08 & 0.11
& 0.44 & 0.45 \\

~~~~~\textit{+Gold MM}
& 0.48 & 0.55
& 0.38 & 0.31
& 0.21 & 0.27 & 0.28 & 0.29
& 0.06 & 0.10 & 0.10 & 0.13
& 0.44 & 0.43 \\

~~~~~\textit{+CoT + Gold MM}
& 0.52 & 0.55
& 0.39 & 0.31
& 0.21 & 0.25 & 0.28 & 0.28
& 0.06 & 0.10 & 0.07 & 0.11
& 0.43 & 0.40 \\

\midrule[0.4pt]

Llama 3.3 70B
& 0.44 & 0.51
& 0.32 & 0.28
& 0.22 & 0.23 & 0.31 & 0.27
& 0.10 & 0.08 & 0.12 & 0.10
& 0.57 & 0.45 \\

~~~~~\textit{+CoT}
& 0.49 & 0.44
& 0.38 & 0.29
& 0.21 & 0.24 & 0.29 & 0.27
& 0.08 & 0.07 & 0.10 & 0.09
& 0.40 & 0.43 \\

~~~~~\textit{+Pred. MM}
& 0.48 & 0.51
& 0.33 & 0.30
& 0.26 & 0.25 & 0.34 & 0.27
& 0.12 & 0.09 & 0.14 & 0.11
& 0.56 & \textbf{0.57} \\

~~~~~\textit{+Gold MM}
& 0.51 & 0.51
& 0.28 & 0.30
& 0.22 & 0.25 & 0.30 & 0.29
& 0.10 & 0.10 & 0.12 & 0.13
& 0.46 & 0.56 \\

~~~~~\textit{+CoT + Gold MM}
& 0.51 & 0.48
& 0.40 & 0.31
& 0.21 & 0.24 & 0.29 & 0.27
& 0.08 & 0.08 & 0.10 & 0.11
& 0.35 & 0.29 \\

\midrule[0.4pt]

GPT-5.5
& 0.56 & 0.59
& 0.44 & 0.35
& 0.24 & 0.32 & 0.33 & 0.36
& 0.09 & \textbf{0.19} & 0.11 & 0.22
& 0.55 & 0.43 \\

~~~~~\textit{+CoT}
& 0.58 & 0.59
& \textbf{0.48} & 0.37
& 0.25 & 0.29 & 0.33 & 0.33
& 0.08 & 0.15 & 0.10 & 0.18
& 0.59 & 0.42 \\

~~~~~\textit{+Pred. MM}
& 0.58 & 0.60
& 0.44 & 0.38
& \textbf{0.31} & 0.33 & \textbf{0.39} & 0.36
& \textbf{0.14} & 0.18 & 0.17 & 0.22
& 0.55 & 0.48 \\

~~~~~\textit{+Gold MM}
& 0.58 & \textbf{0.61}
& 0.46 & 0.37
& 0.25 & 0.17 & 0.33 & \textbf{0.38}
& 0.09 & 0.17 & 0.11 & \textbf{0.25}
& 0.55 & 0.47 \\

~~~~~\textit{+CoT + Gold MM}
& \textbf{0.61} & \textbf{0.61}
& 0.47 & \textbf{0.40}
& 0.25 & 0.34 & 0.33 & 0.36
& 0.08 & \textbf{0.19} & \textbf{0.22} & 0.10
& \textbf{0.60} & 0.48 \\

\midrule[0.4pt]

Claude 4.6 Sonnet
& 0.47 & 0.54
& 0.36 & 0.30
& 0.23 & 0.29 & 0.31 & 0.31
& 0.08 & 0.15 & 0.10 & 0.16
& 0.45 & 0.44 \\

~~~~~\textit{+CoT}
& 0.56 & 0.57
& 0.43 & 0.33
& 0.23 & 0.28 & 0.30 & 0.31
& 0.07 & 0.13 & 0.08 & 0.16
& 0.45 & 0.43 \\

~~~~~\textit{+Pred. MM}
& 0.60 & 0.53
& 0.39 & 0.35
& 0.29 & 0.29 & 0.37 & 0.32
& 0.13 & 0.13 & 0.16 & 0.15
& 0.42 & 0.38 \\

~~~~~\textit{+Gold MM}
& 0.51 & 0.55
& 0.39 & 0.31
& 0.23 & 0.31 & 0.31 & 0.33
& 0.08 & 0.17 & 0.10 & 0.18
& 0.43 & 0.53 \\

~~~~~\textit{+CoT + Gold MM}
& 0.56 & 0.59
& 0.41 & 0.34
& 0.23 & 0.32 & 0.31 & 0.34
& 0.07 & 0.15 & 0.08 & 0.18
& 0.41 & 0.49 \\

\midrule[0.4pt]

Qwen3-4B Fine-tuned
& 0.56 & 0.54
& 0.37 & 0.31
& 0.21 & 0.35 & 0.23 & 0.23
& 0.07 & 0.11 & 0.10 & 0.06
& 0.47 & 0.44 \\

Qwen3-30B-A3B Fine-tuned
& 0.52 & 0.52
& 0.30 & 0.27
& 0.20 & \textbf{0.37} & 0.22 & 0.26
& 0.06 & 0.09 & 0.06 & 0.06
& 0.54 & 0.55 \\

\bottomrule
\end{tabular}}
\caption{
Follower Action Type Accuracy and Recall, Message SBERT and ROUGE for
Guide (G) and Follower (F) roles, and Follower Drawing Accuracy across
six models under $C_{visible}$ and $C_{not\_visible}$.
\textit{Pred. MM} denotes our two-stage ReAct-inspired baseline, which
first predicts the participant's mental model from the observed interaction
trajectory and then conditions next-action prediction on the predicted
mental model. \textit{Gold MM} denotes conditioning on human-annotated
mental models. \textbf{Bolded} numbers indicate the best performance for
each role and canvas visibility condition.
}
\label{tab:next_action_full}
\end{table*}

\begin{table*}[t]
\centering
\small
\resizebox{1.00\linewidth}{!}{
\begin{tabular}{l|cccc|cccc|cccc|cccc|cccc}
\toprule
\multirow{3}{*}{\centering\textbf{Model}}
& \multicolumn{4}{c|}{$\boldsymbol{Accuracy_{\text{Team\_Goal}}}$}
& \multicolumn{4}{c|}{$\boldsymbol{Accuracy_{\text{Partner\_Intent}}}$}
& \multicolumn{4}{c|}{$\boldsymbol{Accuracy_{\text{Self\_Reasoning}}}$}
& \multicolumn{4}{c|}{$\boldsymbol{SBERT_{\text{Rationale}}}$}
& \multicolumn{4}{c}
{$\boldsymbol{ROUGE_{\text{Rationale}}}$} \\
& \multicolumn{2}{c}{$C_{visible}$} & \multicolumn{2}{c|}{$C_{not\_visible}$}
& \multicolumn{2}{c}{$C_{visible}$} & \multicolumn{2}{c|}{$C_{not\_visible}$}
& \multicolumn{2}{c}{$C_{visible}$} & \multicolumn{2}{c|}{$C_{not\_visible}$}
& \multicolumn{2}{c}{$C_{visible}$} & \multicolumn{2}{c|}{$C_{not\_visible}$}
& \multicolumn{2}{c}{$C_{visible}$} & \multicolumn{2}{c}{$C_{not\_visible}$} \\
& G & F & G & F
& G & F & G & F
& G & F & G & F
& G & F & G & F
& G & F & G & F \\
\midrule
Qwen3-35B-A3B            & \textbf{0.48} & 0.69 & 0.41 & 0.64 & \textbf{0.58} & 0.73 & 0.38 & 0.67 & \textbf{0.31} & 0.56 & 0.32 & 0.52 & \textbf{0.41} & 0.46 & 0.44 & 0.52 & 0.14 & 0.20 & 0.15 & 0.20 \\
Llama 3.3 70B            & 0.43 & 0.71 & \textbf{0.56} & 0.64 & 0.38 & 0.76 & 0.34 & 0.72 & 0.23 & 0.55 & 0.31 & 0.53 & \textbf{0.41} & 0.47 & 0.44 & 0.54 & 0.17& 0.21& \textbf{0.18} & 0.26\\
GPT-5.5                  & 0.41 & 0.72 & 0.35 & 0.68 & 0.37 & 0.75 & 0.36 & 0.70 & 0.29 & 0.60 & 0.32 & 0.59 & 0.40 & 0.51 & \textbf{0.45} & 0.55 & 0.16 &0.24 & 0.17& 0.24\\
Claude 4.6 Sonnet        & \textbf{0.48} & 0.75 & 0.45 & 0.68 & 0.51 & 0.76 & 0.45 & 0.71 & 0.27 & 0.56 & 0.30 & 0.55 & \textbf{0.41} & 0.52 & \textbf{0.45} & 0.55 & 0.15& 0.25 & 0.16 & 0.21 \\
Qwen3-4B Fine-tuned      & 0.37 & \textbf{0.81} & 0.51 & \textbf{0.88} & 0.40 & \textbf{0.84} & \textbf{0.47} & \textbf{0.84} & 0.28 & \textbf{0.65} & 0.30 & \textbf{0.70} & 0.37 & \textbf{0.76} & 0.33 & 0.64 & \textbf{0.18} & \textbf{0.68} & 0.16 & 0.46 \\
Qwen3-30B-A3B Fine-tuned & 0.47 & 0.55 & 0.39 & 0.55 & 0.38 & 0.78 & 0.46 & 0.77 & 0.29 & 0.54 & \textbf{0.33} & 0.54 & 0.34 & 0.61 & 0.38 & \textbf{0.66} & 0.17 & 0.44 & \textbf{0.18} & \textbf{0.55} \\
\bottomrule
\end{tabular}}
\caption{Team Goal Accuracy, Partner Intent Accuracy, Self-Reasoning Accuracy, Rationale SBERT, and Rationale ROUGE across six models in Guide (G) and Follower (F) roles under $C_{visible}$ and $C_{not\_visible}$. \textbf{Bolded} numbers indicate the best performance for each role and canvas visibility condition.}
\label{tab:mental_model_full}
\end{table*}

\section{Prompts}
\label{app:prompts}
\subsection{Next Action Prediction}
\
\newtcolorbox{promptbox}[1][]{
  breakable,
  colback=gray!12,
  colframe=gray!50,
  boxrule=0.8pt,
  arc=2pt,
  title=#1,
  fonttitle=\bfseries\small,
  coltitle=black,
  colbacktitle=gray!30,
  left=6pt, right=6pt, top=2pt, bottom=2pt,
  fontupper=\small
  before skip=0pt
}

\begin{promptbox}[Follower's Prompt]
\textbf{<Task Description>} \\
You are participating in a two-player collaborative map-reproduction task. There are two roles: a guide and a follower. The guide can see a map with all landmarks and the correct route. The follower can see a similar map with all landmarks but without the route. The two players need to communicate and coordinate so that the follower can reproduce the guide’s route on the follower’s map.

The task unfolds through a sequence of actions. At each step, the follower may send a message, draw part of the route, erase part of the route, undo the latest edit, or reset the drawing. \\[2pt]

\textbf{<Role Description>} \\
You are role-playing the follower in this task. Your goal is not to solve the task perfectly, but to authentically simulate what this specific human follower would most likely do next, given their persona and the previous interaction history. \\[2pt]

\textbf{<Game Rules>} \\
- You and the guide cannot directly see each other’s maps. \\
- One landmark on your map is misplaced compared with the guide’s map. However, you should follow the guide's instructions and reproduce the route on your map. \\
- You may need to ask clarification questions, acknowledge instructions, draw based on your current understanding, correct previous drawing errors, or wait for more guidance.\\[2pt]

\textbf{<Participant context>} \\
Participant role: \texttt{PARTICIPANT\_ROLE} \\
Participant name: \texttt{PARTICIPANT\_NAME} \\

\texttt{\{COLLABORATION\_PROFILE\}}\\

Use this profile as a soft behavioral tendency, not a fixed rule. The next mental model should still be primarily grounded in the interaction history and current map state. If the history shows a different behavioral pattern, prioritize the observed interaction history over the TeamQ profile. \\[2pt]

\textbf{<Action Space>} \\
You must choose exactly one next action from the following action types:\\
\textbf{message}: Send a message to communicate with the guide.\\
\textbf{draw}: Draw a route segment on the follower’s map. The content must be an ordered list of [row, col] cells in the direction of travel.\\
\textbf{erase}: Erase part of the current drawing. The content must be an ordered list of [row, col] cells to erase. \\
\textbf{undo}: Undo latest route edit.\\
\textbf{reset}: Clear entire drawing.\\[2pt]

\textbf{<Map Interpretation>} \\
Discrete grid, 0-based [row, col]. Origin top-left [0, 0]; row increases downward, col increases rightward.
Any landmark with "type": "blocked" has a "cells" list, which means those cells are impassable. Your route must NEVER include them.
Use bbox / centroid from the landmark reference plus the map image to locate named landmarks. "Bottom / top / left / right" of a landmark refers to that region of the landmark, not the whole map. Paths to a landmark corner usually require BOTH row and col to change — not a single long horizontal or vertical segment.\\[2pt]

\textbf{<Current Map>} \\
\texttt{\{CURRENT\_MAP\}}
\\[2pt]

\textbf{<Response Format>} \\
\{{ \\
\hspace*{1em}"action\_type": $"message | draw | erase | undo | reset"$, \\
\hspace*{1em}"action\_content": $"..."$, \\
\hspace*{1em}"rationale": $"..."$ \\
}\}
\\

For action\_content: \\
- If action\_type is "message", action\_content must be the exact message text the follower would send. \\
- If action\_type is "draw" or "erase", action\_content must be an ordered list of [row, col] cells, for example: [[12, 8], [12, 9], [13, 9]]. \\
- If action\_type is "undo" or "reset", action\_content must be an empty string "". \\
\\
For rationale:\\
Briefly explain why this action is the most likely next action for this follower, based on the persona and interaction history. The rationale should be concise and should not introduce information that is not visible in the input.\\[2pt]

\textbf{<Instructions for aligning with human behaviors>} \\
- Given the interaction history, predict the single next action that this follower would most likely take. Your prediction should be grounded in: \\
\hspace*{1em}- the guide’s most recent messages;\\
\hspace*{1em}- the follower’s previous actions and communication style; \\
\hspace*{1em}- the follower’s persona.\\
- Do not predict an ideal or optimal action unless it is also likely for this specific follower. Human participants may be incomplete, cautious, redundant, informal, uncertain, or locally focused. Preserve these behavioral patterns when they appear in the history.\\
- Do not add unnecessary politeness, formal language, or overly detailed explanations unless this follower has shown that style.\\
- Do not invent information that is not supported by the interaction history.\\
- Do not mention that you are an AI, a simulator, or making a prediction. \\[2pt]

\textbf{<Interaction History>} \\
\texttt{\{INTERACTION\_HISTORY\}}
\end{promptbox}

\begin{promptbox}[Guide's Prompt]
\textbf{<Task Description>} \\
You are participating in a two-player collaborative map-reproduction task. There are two roles: a guide and a follower. The guide can see a map with all landmarks and the correct route. The follower can see a similar map with all landmarks but without the route. The two players need to communicate and coordinate so that the follower can reproduce the guide’s route on the follower’s map.

The task unfolds through a sequence of actions. At each step, the guide can only send a message. \\[2pt]

\textbf{<Role Description>} \\
You are role-playing the guide in this task. Your goal is not to solve the task perfectly, but to authentically simulate what this specific human guide would most likely do next, given their persona and the previous interaction history. \\[2pt]

\textbf{<Game Rules>} \\
- You and the follower cannot directly see each other’s maps. \\
- One landmark on the follower’s map is misplaced compared with your map. You should give instructions so the follower can reproduce the route on their map. \\
- You may need to ask clarification questions, acknowledge the follower’s messages, give route directions based on your understanding of their progress, correct or refine earlier instructions, or wait for more information from the follower.\\[2pt]

\textbf{<Participant context>} \\
Participant role: \texttt{PARTICIPANT\_ROLE} \\
Participant name: \texttt{PARTICIPANT\_NAME} \\

\texttt{\{COLLABORATION\_PROFILE\}}\\

Use this profile as a soft behavioral tendency, not a fixed rule. The next mental model should still be primarily grounded in the interaction history and current map state. If the history shows a different behavioral pattern, prioritize the observed interaction history over the TeamQ profile. \\[2pt]

\textbf{<Action Space>} \\
The only action type is:\\
\textbf{message}: Send a message to communicate with the follower.\\[2pt]

\textbf{<Map Interpretation>} \\
Discrete grid, 0-based [row, col]. Origin top-left [0, 0]; row increases downward, col increases rightward.
Any landmark with "type": "blocked" has a "cells" list, which means those cells are impassable. Your route must NEVER include them.
Use bbox / centroid from the landmark reference plus the map image to locate named landmarks. "Bottom / top / left / right" of a landmark refers to that region of the landmark, not the whole map. Paths to a landmark corner usually require BOTH row and col to change — not a single long horizontal or vertical segment.\\[2pt]

\textbf{<Current Map>} \\
\texttt{\{CURRENT\_MAP\}}
\\[2pt]

\textbf{<Response Format>} \\
\{{ \\
\hspace*{1em}"action\_type": $"message"$, \\
\hspace*{1em}"action\_content": $"..."$, \\
\hspace*{1em}"rationale": $"..."$ \\
}\}
\\

For action\_content: \\
- If action\_type is "message", action\_content must be the exact message text the guide would send. \\
\\
For rationale:\\
Briefly explain why this action is the most likely next action for this guide, based on the persona and interaction history. The rationale should be concise and should not introduce information that is not visible in the input.\\[2pt]

\textbf{<Instructions for aligning with human behaviors>} \\
- Given the interaction history, predict the single next action that this guide would most likely take. Your prediction should be grounded in: \\
\hspace*{1em}- the follower’s most recent actions;\\
\hspace*{1em}- the guide’s previous actions and communication style; \\
\hspace*{1em}- the guide’s persona.\\
- Do not predict an ideal or optimal action unless it is also likely for this specific guide. Human participants may be incomplete, cautious, redundant, informal, uncertain, or locally focused. Preserve these behavioral patterns when they appear in the history.\\
- Do not add unnecessary politeness, formal language, or overly detailed explanations unless this follower has shown that style.\\
- Do not invent information that is not supported by the interaction history.\\
- Do not mention that you are an AI, a simulator, or making a prediction. \\[2pt]

\textbf{<Interaction History>} \\
\texttt{\{INTERACTION\_HISTORY\}}
\end{promptbox}

\subsection{Mental Model Prediction}

\begin{promptbox}[Follower's Prompt]
\textbf{<Task Description>} \\
You are simulating a human participant’s action-level mental model in a two-player collaborative map-reproduction task. In this task, there are two roles: guide and follower. The guide can see a map with all landmarks and the correct route. The follower can see a similar map with all landmarks but does not see the route. The two players need to communicate and coordinate so that the follower can reproduce the guide’s route on the follower’s map. The task unfolds as a sequence of actions. At each step, the follower may send a message, draw part of the route, erase part of the route, undo the latest route edit, or reset the drawing. Your task is to predict the follower’s mental model at the current action moment. \\[2pt]

\textbf{<Role Description>} \\
You are simulating the follower’s mental model during the map task. Given the participant’s persona, the interaction history, the current action, the current drawing state, and any previous mental model annotations, predict what this specific follower would most likely report about: \\

1. what the team was trying to do; \\
2. what they thought the guide was trying to do; \\
3. what they themselves were trying to do; \\
4. why they took or understood the current action in that way. \\

The prediction should reflect the follower’s subjective understanding at that moment, not the objective ground truth of the task. \\[2pt]

\textbf{<Game Rules>} \\
- The follower and the guide cannot directly see each other’s maps. \\
- One landmark on the follower’s map is misplaced compared with the guide’s map. However, the follower should follow the guide's instructions and reproduce the route on the follower's map.\\[2pt]

\textbf{<Participant context>} \\
Participant role: \texttt{PARTICIPANT\_ROLE} \\
Participant name: \texttt{PARTICIPANT\_NAME} \\

\texttt{\{COLLABORATION\_PROFILE\}}\\

Use this profile as a soft behavioral tendency, not a fixed rule. The next mental model should still be primarily grounded in the interaction history and current map state. If the history shows a different behavioral pattern, prioritize the observed interaction history over the TeamQ profile. \\[2pt]

\textbf{<Mental Model Annotation Task>} \\
The participant provided mental model annotations after completing the Map Task. For each action, they were asked to recall what they were thinking at that specific moment. You need to predict the participant’s annotation for the current action. The annotation contains four fields: \\

1. team\_goal \\
What the follower thought the team was trying to do at that moment. Choose exactly one label: \\ 
- "Still figuring out what we needed to do"
- "Working toward a shared understanding"
- "Clear on what to do and working on it"
- "Something was unclear and we were working it out"
- "Other" \\

2. partner\_intent \\
What the follower thought the guide was trying to do or understood at that moment. Choose exactly one label:\\
- "Understood the situation and we were on the same page"
- "Probably understood our situation but I was not fully sure"
- "Is waiting for more information to understand the situation"
- "Misunderstood and we were not aligned"
- "Gave no clear signal either way"
- "Other" \\

3. self\_reasoning \\
What the follower thought they themselves were trying to do at that moment. Choose exactly one label: \\
- "Executing a plan we already agreed on"
- "Exploring on my own to gather information"
- "Confirming the situation with my partner"
- "Grounding by sharing or requesting information to align"
- "Repairing a mistake or misunderstanding"
- "Waiting for more information"
- "Other" \\

If you select 'Other' for any label, you must provide a specific, meaningful label to replace 'Other'—do not just leave it as 'Other'. \\

4. rationale \\
A short free-form explanation, written from the follower’s perspective, describing what the team, the guide, and the follower were trying to do at that action moment. The rationale should sound like the participant’s own retrospective explanation, not an external analysis. \\

\textbf{<Map Interpretation>} \\
Discrete grid, 0-based [row, col]. Origin top-left [0, 0]; row increases downward, col increases rightward.
Any landmark with "type": "blocked" has a "cells" list, which means those cells are impassable. Your route must NEVER include them.
Use bbox / centroid from the landmark reference plus the map image to locate named landmarks. "Bottom / top / left / right" of a landmark refers to that region of the landmark, not the whole map. Paths to a landmark corner usually require BOTH row and col to change — not a single long horizontal or vertical segment.\\[2pt]

\textbf{<Current Map>} \\
\texttt{\{CURRENT\_MAP\}}
\\[2pt]

\textbf{<Current follower action (you are simulating the mental model behind this action)>} \\
\texttt{\{CURRENT\_ACTION\}}
\\[2pt]

\textbf{<Response Format>} \\
Return only a valid JSON object. Do not include markdown, explanations, or extra text outside the JSON. The JSON must have exactly the following fields: \\

\{{\\
\hspace*{1em}"team\_goal": $"..."$, \\
\hspace*{1em}"partner\_intent": $"..."$, \\
\hspace*{1em}"self\_reasoning": $"..."$, \\
\hspace*{1em}"rationale": $"..."$ \\
}\}
\\
The three label fields must exactly match one of the allowed labels. \\[2pt]

\textbf{<Instructions for aligning with human behaviors>} \\
- Given the interaction history and previous mental models, predict this follower's mental model for the current action. Your prediction should be grounded in: \\
\hspace*{1em}- the guide’s most recent messages;\\
\hspace*{1em}- the follower’s previous actions and communication style; \\
\hspace*{1em}- the follower’s previous mental models; \\
\hspace*{1em}- the follower’s persona.\\
- Follow the participant's previous behaviors/habits (e.g., writing styles, preferences, etc.) when reporting their mental model. \\
- Do not add unnecessary politeness, formal language, or overly detailed explanations unless this follower has shown that style.\\ 
- Write the rationale in a first-person perspective, as if the participant is recalling their own thought process after the task.\\
- Do not invent information that is not supported by the interaction history.\\
- Do not mention that you are an AI, a simulator, or making a prediction. \\[2pt]

\textbf{<Interaction History>} \\
\texttt{\{INTERACTION\_HISTORY\}}
\end{promptbox}

\begin{promptbox}[Guide's Prompt]
\textbf{<Task Description>} \\
You are simulating a human participant’s action-level mental model in a two-player collaborative map-reproduction task. In this task, there are two roles: guide and follower. The guide can see a map with all landmarks and the correct route. The follower can see a similar map with all landmarks but does not see the route. The two players need to communicate and coordinate so that the follower can reproduce the guide’s route on the follower’s map. The task unfolds as a sequence of actions. At each step, the guide will send a message. Your task is to predict the guide’s mental model at the current action moment.\\[2pt]

\textbf{<Role Description>} \\
You are simulating the guide’s mental model during the map task. Given the participant’s persona, the interaction history, the current action, and any previous mental model annotations, predict what this specific guide would most likely report about: \\

1. what the team was trying to do; \\
2. what they thought the guide was trying to do; \\
3. what they themselves were trying to do; \\
4. why they took or understood the current action in that way. \\

The prediction should reflect the guide’s subjective understanding at that moment, not the objective ground truth of the task.\\[2pt]

\textbf{<Game Rules>} \\
- The follower and the guide cannot directly see each other’s maps. \\
- One landmark on the follower’s map is misplaced compared with the guide’s map. However, the follower should follow the guide's instructions and reproduce the route on the follower's map.\\[2pt]

\textbf{<Participant context>} \\
Participant role: \texttt{PARTICIPANT\_ROLE} \\
Participant name: \texttt{PARTICIPANT\_NAME} \\

\texttt{\{COLLABORATION\_PROFILE\}}\\

Use this profile as a soft behavioral tendency, not a fixed rule. The next mental model should still be primarily grounded in the interaction history and current map state. If the history shows a different behavioral pattern, prioritize the observed interaction history over the TeamQ profile. \\[2pt]

\textbf{<Mental Model Annotation Task>} \\
The participant provided mental model annotations after completing the Map Task. For each action, they were asked to recall what they were thinking at that specific moment. You need to predict the participant’s annotation for the current action. The annotation contains four fields: \\

1. team\_goal \\
What the guide thought the team was trying to do at that moment. Choose exactly one label: \\ 
- "Still figuring out what we needed to do"
- "Working toward a shared understanding"
- "Clear on what to do and working on it"
- "Something was unclear and we were working it out"
- "Other" \\

2. partner\_intent \\
What the guide thought the guide was trying to do or understood at that moment. Choose exactly one label:\\
- "Understood the situation and we were on the same page"
- "Probably understood our situation but I was not fully sure"
- "Is waiting for more information to understand the situation"
- "Misunderstood and we were not aligned"
- "Gave no clear signal either way"
- "Other" \\

3. self\_reasoning \\
What the guide thought they themselves were trying to do at that moment. Choose exactly one label: \\
- "Executing a plan we already agreed on"
- "Exploring on my own to gather information"
- "Confirming the situation with my partner"
- "Grounding by sharing or requesting information to align"
- "Repairing a mistake or misunderstanding"
- "Waiting for more information"
- "Other" \\

If you select 'Other' for any label, you must provide a specific, meaningful label to replace 'Other'—do not just leave it as 'Other'. \\

4. rationale \\
A short free-form explanation, written from the follower’s perspective, describing what the team, the guide, and the follower were trying to do at that action moment. The rationale should sound like the participant’s own retrospective explanation, not an external analysis. \\

\textbf{<Map Interpretation>} \\
Discrete grid, 0-based [row, col]. Origin top-left [0, 0]; row increases downward, col increases rightward.
Any landmark with "type": "blocked" has a "cells" list, which means those cells are impassable. Your route must NEVER include them.
Use bbox / centroid from the landmark reference plus the map image to locate named landmarks. "Bottom / top / left / right" of a landmark refers to that region of the landmark, not the whole map. Paths to a landmark corner usually require BOTH row and col to change — not a single long horizontal or vertical segment.\\[2pt]

\textbf{<Current Map>} \\
\texttt{\{CURRENT\_MAP\}}
\\[2pt]

\textbf{<Current guide action (you are simulating the mental model behind this action)>} \\
\texttt{\{CURRENT\_ACTION\}}
\\[2pt]

\textbf{<Response Format>} \\
Return only a valid JSON object. Do not include markdown, explanations, or extra text outside the JSON. The JSON must have exactly the following fields: \\

\{{\\
\hspace*{1em}"team\_goal": $"..."$, \\
\hspace*{1em}"partner\_intent": $"..."$, \\
\hspace*{1em}"self\_reasoning": $"..."$, \\
\hspace*{1em}"rationale": $"..."$ \\
}\}
\\
The three label fields must exactly match one of the allowed labels. \\[2pt]

\textbf{<Instructions for aligning with human behaviors>} \\
- Given the interaction history and previous mental models, predict this guide's mental model for the current action. Your prediction should be grounded in: \\
\hspace*{1em}- the follower’s most recent actions;\\
\hspace*{1em}- the guide’s previous actions and communication style; \\
\hspace*{1em}- the guide’s previous mental models; \\
\hspace*{1em}- the guide’s persona.\\
- Follow the participant's previous behaviors/habits (e.g., writing styles, preferences, etc.) when reporting their mental model. \\
- Do not add unnecessary politeness, formal language, or overly detailed explanations unless this follower has shown that style.\\ 
- Write the rationale in a first-person perspective, as if the participant is recalling their own thought process after the task.\\
- Do not invent information that is not supported by the interaction history.\\
- Do not mention that you are an AI, a simulator, or making a prediction. \\[2pt]

\textbf{<Interaction History>} \\
\texttt{\{INTERACTION\_HISTORY\}}
\end{promptbox}

\end{document}